\documentclass{article} % For LaTeX2e
\usepackage{iclr2024_conference,times}

\usepackage{amsmath,amsfonts,bm}

\def\eqref#1{equation~\ref{#1}}
\def\1{\bm{1}}

\def\mS{{\bm{S}}}
\def\mT{{\bm{T}}}

\DeclareMathAlphabet{\mathsfit}{\encodingdefault}{\sfdefault}{m}{sl}
\SetMathAlphabet{\mathsfit}{bold}{\encodingdefault}{\sfdefault}{bx}{n}

\usepackage{hyperref}
\usepackage{url}
\usepackage{times}
\usepackage{epsfig}
\usepackage{graphicx}
\usepackage{amsmath}
\usepackage{amssymb}
\usepackage[linesnumbered,ruled,vlined]{algorithm2e}
\usepackage{booktabs}
\usepackage{multirow}
\usepackage{subfigure}
\usepackage{paralist}
\usepackage{tabularx}
\usepackage{array}
\iclrfinalcopy
\usepackage{amsmath}

\title{Dataset Distillation via Adversarial Prediction Matching}

\author{Mingyang Chen\textsuperscript{1,2} Bo Huang\textsuperscript{1,2} Junda Lu\textsuperscript{3} Bing Li\textsuperscript{4} Yi Wang\textsuperscript{5} Minhao Cheng\textsuperscript{2} Wei Wang\textsuperscript{1}  \\
\textsuperscript{1}The Hong Kong University of Science and Technology (Guangzhou)  \\
\textsuperscript{2}The Hong Kong University of Science and Technology \textsuperscript{3}CSIRO Data61, Australia\\
\textsuperscript{4}University of Electronic Science and Technology of China \textsuperscript{5}Dongguan University of Technology \\
\texttt{\{mchenbt,bhuangas\}@connect.ust.hk}, \texttt{Junda.Lu@data61.csiro.au}\\ 
\texttt{indof47@gmail.com}, \texttt{wangyi@dgut.edu.cn}, \texttt{\{minhaocheng,weiwcs\}@ust.hk}      \\
}
\begin{document}

\maketitle

\begin{abstract}

Dataset distillation is the technique of synthesizing smaller condensed datasets from large original datasets while retaining necessary information to persist the effect. In this paper, we approach the dataset distillation problem from a novel perspective: we regard minimizing the prediction discrepancy on the real data distribution between models, which are respectively trained on the large original dataset and on the small distilled dataset, as a conduit for condensing information from the raw data into the distilled version. An adversarial framework is proposed to solve the problem efficiently. In contrast to existing distillation methods involving nested optimization or long-range gradient unrolling,  our approach hinges on single-level optimization. 
This ensures the memory efficiency of our method and provides a flexible tradeoff between time and memory budgets, allowing us to distil ImageNet-1K using a minimum of only 6.5GB of GPU memory. Under the optimal tradeoff strategy, it requires only 2.5$\times$ less memory and 5$\times$ less runtime compared to the state-of-the-art.
Empirically, our method can produce synthetic datasets just 10\% the size of the original, yet achieve, on average, 94\% of the test accuracy of models trained on the full original datasets including ImageNet-1K, significantly surpassing state-of-the-art. Additionally, extensive tests reveal that our distilled datasets excel in cross-architecture generalization capabilities. Our code is available at \url{https://github.com/mchen725/DD_APM/}.
\end{abstract}
\section{Introduction}
%The astonishing innovations driven by deep learning techniques in various filed is inevitably relay on the significant data explosion in the past decades. A large-scale dataset is generally regarded as a pivotal component for achieving state-of-the-art performances in the deep learning era, which, nevertheless, requires severe computational resources for the model adjustment tasks such as hyperparameter tuning~\citep{} or architecture search~\citep{}. \\
% The advancements in various fields driven by deep learning techniques are inevitably reliant on the significant data explosion of the past decades. 
Cutting-edge machine learning models across diverse domains are progressively dependent on extensive datasets. Nevertheless, the enormity of these datasets introduces formidable challenges in aspects such as data storage, preprocessing, model training, continual learning~\citep{DBLP:conf/nips/AljundiBTCCLP19,DBLP:conf/icml/ZhaoB21}, and other auxiliary tasks including hyperparameter tuning~\citep{DBLP:conf/nips/BergstraBBK11,DBLP:conf/nips/BaikCCKL20} and architecture search~\citep{DBLP:journals/corr/abs-1802-03268,DBLP:conf/kdd/JinSH19}.
Exploring surrogate datasets emerges as a viable solution, offering comparable efficacy while significantly reducing scale. While existing research on core-set selection has underscored the feasibility of employing representative subsets from the original datasets~\citep{DBLP:conf/uai/ChenWS10,DBLP:conf/cvpr/RebuffiKSL17,DBLP:conf/nips/AljundiLGB19,DBLP:conf/iclr/TonevaSCTBG19}, Dataset Distillation (DD) is a nascent approach that generate \textit{new} data instances not present in the original sets. This methodology notably expands the scope for optimization, potentially retaining optimal training signals from the principal dataset, and has demonstrated remarkable efficacy and effectiveness \footnote{https://github.com/Guang000/Awesome-Dataset-Distillation}.

Dataset distillation via imitating the behaviour of ``teacher models'' acquired through training on original datasets, by the intermediary ``student models'' trained on distilled datasets is perceived as an effective solution \citep{DBLP:conf/icml/ZhaoB21,DBLP:conf/cvpr/Cazenavette00EZ22b}. These methods utilize the teacher-student paradigm as a medium to condense knowledge from raw data into a more succinct form. Particularly, Matching Training Trajectory (MTT)  frameworks \citep{DBLP:conf/cvpr/DuJTZ023,DBLP:conf/icml/CuiWSH23} exhibit superior performance over alternate distillation paradigms. Compared to its predecessor, i.e., the gradient matching framework \citep{DBLP:conf/iclr/ZhaoMB21}, its superior performance can be attributed to the strategy of matching long-term training trajectories of teachers rather than short-term gradients. However, over-fitting on the synthetic dataset can cause a remarkable disparity between the obtained and real trajectories, deteriorating the effectiveness and incurring considerable memory burdens during gradient derivation w.r.t synthetic samples. Hence, such methodologies remain confined to a localized imitation of behaviours elicited by teachers. 

\begin{figure}[t]
	\begin{center}
		\centerline
		{\includegraphics[width=0.95\textwidth]{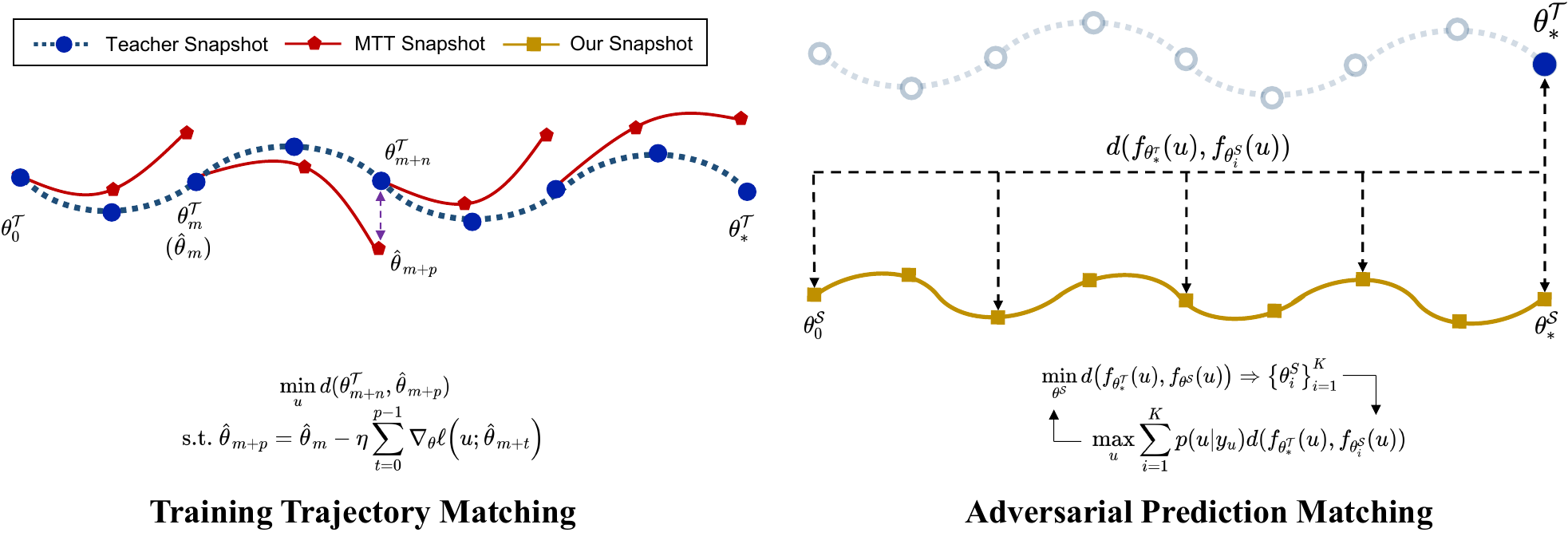}}
		\caption{We contrast our adversarial prediction matching framework with the training trajectory matching (MTT) framework~\citep{DBLP:conf/cvpr/Cazenavette00EZ22b}. Our method has three primary advantages: \textit{1.} It imitates the converged teachers' prediction, avoiding short-sightedness on local trajectories; \textit{2.} Synthetic samples $u$ are updated by a single-level loss function, significantly enhancing memory complexity; \textit{3.} It requires only one well-trained teacher, markedly diminishing storage overhead.}
		\label{fig:outline}
	\end{center}
	
\end{figure}

Prompted by these insights, we formally formulate a novel prediction-matching-based objective. This objective motivates synthetic data to enable student models to directly emulate the predictions of teacher models on the authentic data distributions of the original datasets, circumventing the risk of the distilled features becoming shortsighted. To effectively optimize the objective and bypass the nested optimization challenges in various prevailing methods \citep{DBLP:conf/cvpr/DuJTZ023,DBLP:conf/icml/LooHLR23,DBLP:conf/icml/CuiWSH23}, we formulate an adversarial framework. A depiction of the basic proposition of our adversarial framework and its superiorities over the trajectory matching paradigm are provided in Figure \ref{fig:outline}. Intuitively, we obtain training snapshots of students by aligning predictions with teachers' on synthetic samples, then adversarially refine these samples to approach ``critical points'' in the real data distribution, which can cause substantial prediction disagreement between teachers and the aggregated snapshots with high probabilities in the original distribution. This strategy adeptly enables the distilled data to encapsulate crucial features from the original, motivating consistent alignment between intermediary students and teachers throughout the training phase.
% This strategy efficiently prompts the distilled data to condense essential features from the original, ensuring effective signals for aligning between students and teachers alongside the whole training stage.

Notably, although the utilized teacher-student paradigm in DD has a similar form to Knowledge Distillation (KD) \citep{DBLP:journals/corr/HintonVD15,DBLP:journals/corr/abs-1912-11006}, especially our method hinges a prediction-matching paradigm, the research focal points are orthogonal. Instead of matching predictions between \textbf{two designated models}, the goal of DD is to create synthetic datasets endowed with prominent generalization ability that can off-the-shelf train unseen models. To this end, we devote ourselves to harnessing adequate information for the distilled data from the original, employing the prediction-matching paradigm as a conduit.

The contributions of this paper can be summarized as:
\begin{itemize} 
\item A novel imitation-based prediction-matching notation is introduced for effective knowledge condensation from original datasets to distilled counterparts by aligning predictions of models, trained respectively on original and distilled datasets, over the authentic data distribution. An adversarial framework is consequently proposed to formulate the problem as a min-max game, allowing for trivial single-level optimization of distilled samples.

\item Our method is proven to exhibit low memory complexity, offering a flexible tradeoff between memory consumption and runtime. Empirically, it enables the distillation of the ImageNet-1K dataset using a mere 6.5GB of GPU memory, achieving superior outcomes with only 2.5$\times$ less memory and 5$\times$ reduced runtime compared to the foremost approach.

\item Experimental results illustrate that our distilled datasets, with just 10\% the size of the original, can averagely attain 94\% of the test accuracies of models than training on the original benchmark datasets, significantly surpassing existing solutions. Additionally, the distilled dataset demonstrates superior performance in cross-architecture generalization and excels in the Neural Architecture Search (NAS) task.
\end{itemize}

\section{Related Work of Dataset Distillation}

\textbf{Solving Bi-Level Optimizations.\quad}A common approach in dataset distillation is to model the task as a bi-level optimization problem, aiming to minimize the generalization error on the original data caused by models trained on the distilled data~\citep{DBLP:journals/corr/abs-1811-10959,DBLP:journals/corr/abs-2006-08572,zhou2022dataset,DBLP:conf/iclr/NguyenCL21,DBLP:conf/icml/ZhaoB21,DBLP:conf/icml/LooHLR23}. However, solving this problem is non-trivial, especially when the employed proxy models are optimized by gradient descent because it requires unrolling a complicated computational graph. To tackle this challenge, several recent works propose to approximate the model training with Kernel Ridge Regression (KRR) to get a closed-form solution for the optimum weights~\citep{zhou2022dataset,DBLP:conf/iclr/NguyenCL21} or directly solving the problem by the Implicit Gradient (IG) through convexifying the model training by the Neural Tangent Kernel (NTK)~\citep{DBLP:conf/icml/LooHLR23}. Nonetheless, these methods either require extensive computations or are compromised by the looseness in the convex relaxation. %

\textbf{Imitation-based Approaches.\quad}Another widespread notion in dataset distillation involves mimicking specific behaviours elicited by the original dataset with those of the distilled dataset. Following this concept, the gradient matching frameworks \citep{DBLP:conf/iclr/ZhaoMB21,DBLP:conf/icml/ZhaoB21,DBLP:conf/icml/KimKOYSJ0S22} and training trajectory matching frameworks  \citep{DBLP:conf/cvpr/Cazenavette00EZ22b,DBLP:conf/cvpr/DuJTZ023} propose to emulate the training dynamics of models experienced on the original dataset, making models trained on the distilled data converge to a neighbourhood in the parameter space. Although a significant improvement has been achieved by state-of-the-art methods, the update of synthetic data is typically memory-intensive as it also requires unrolling the computation graph of multistep gradient descent. 
While \citet{DBLP:conf/icml/CuiWSH23} addresses this issue by allowing constant memory usage during gradient derivation, aligning trajectories obtained from extensive steps over synthetic data with long-term real trajectories becomes challenging. Consequent tradeoff also leads to inevitable truncation bias~\citep{zhou2022dataset}. Additionally, distribution matching methods \citep{DBLP:journals/corr/abs-2110-04181,DBLP:conf/cvpr/WangZPZYWHBWY22} heuristically propose to align the feature distribution of original data with that of distilled data output by several proxy extractors,  exhibiting inferior effectiveness. \\

\section{Method}
%Distilling datasets via imitating the behaviour of teacher networks, acquired through training on real datasets, by the intermediary student models trained on these distilled datasets is perceived as an effective solution of the dataset distillation \citep{DBLP:conf/cvpr/Cazenavette00EZ22b,DBLP:conf/icml/ZhaoB21}. These approaches utilize the teacher-student framework as a medium to condense knowledge from raw data to a refined version. Particularly, Matching Training Trajectory (MTT)  frameworks \citep{DBLP:conf/cvpr/DuJTZ023,DBLP:conf/icml/CuiWSH23} achieve superior performance than other distillation frameworks. Compared to its predecessor, i.e. the gradient matching framework \citep{DBLP:conf/iclr/ZhaoMB21}, its superior performance can be attributed to the strategy of matching long-term training trajectories of teachers rather than short-term gradients. However, empirical evidences suggest that over-updating on the synthetic dataset can lead to a remarkable disparity between the obtained and real trajectories, deteriorating the results and imposing substantial memory overheads when deriving gradients w.r.t synthetic samples. Consequently, these frameworks remain confined to local imitation of the behaviours of teacher models derived from original datasets. Motivated by these observations, we propose our prediction matching framework, which generates synthetic data enabling models to directly mimic the predictions of teacher models on the real data distribution of the original dataset to effectively avoid the distilled features being myopic.

\subsection{Dataset Distillation via Matching Prediction on Real Data Distribution}
Suppose a large labelled training dataset $\mathcal{T} = \left\{ (x_i, y_i) \right\}_{i=1}^{|\mathcal{T}|}$, where $x_i \in \mathbb{R}^d\) and \(y_i \in \{1, 2, \ldots, C\}$, is given to be distilled into a smaller distilled dataset $\mathcal{S} = \left\{ (u_i, v_i) \right\}_{i=1}^{|\mathcal{S}|}$, with $u_i \in \mathbb{R}^d$ and $v_i\in\mathbb{R}^C$, such that $|\mathcal{S}| \ll |\mathcal{T}|$.
%Following the common protocol of distilling non-long-tailed classification datasets \citep{}, each 
Our objective is to construct $\mathcal{S}$ such that, for a given model architecture parameterized by $\theta\sim p(\theta)$, models trained respectively on $\mathcal{S}$ and $\mathcal{T}$ have the minimum discrepancy when predicting over the original data distribution with $p(x)= \sum_y p(x | y) p(y)$:
\begin{equation}\label{eq:ini}
	\min _{\mathcal{S}} \sum_{y=1}^{C} \int p(x | y) \cdot d\left(f_{\theta^\mathcal{T}}(x), f_{\theta^\mathcal{S}}(x)\right) d x
\end{equation}
We omit $p\left(y\right)$ as we assume it follows a uniform distribution. Where, $d\left(.,.\right)$ refers to a distance function and  $f_{\theta}\left(.\right)$ refers to the logtis output by the penultimate layer of models before the softmax. We employ logits for the prediction matching instead of the confidences normalized by the softmax function because logits contain more information for knowledge transfer~\citep{yuan2020revisiting,role-kd,yin2020dreaming}. 
$\theta^{\mathcal{T}}$ and $\theta^{\mathcal{S}}$ refer to the optimized parameters obtained by minimizing empirical loss over $\mathcal{T}$ and $\mathcal{S}$ respectively. Typically, this process is performed by an iterative optimization algorithm, denoted as $Alg$, such as gradient descent. For instance, $ \theta^{\mathcal{S}} = Alg(\ell(\theta, \mathcal{S}), E) $, where \( \ell(.,.) \) is any suitable loss function for model training, and \( E \) denotes the number of training epochs.

We emphasize that minimizing the objective in Equation (\ref{eq:ini}) is non-trivial as two reasons:
\textit{i}) calculating the gradient w.r.t $\mathcal{S}$ requires unrolling the recursive computation graph through the entire training process of $\theta^{\mathcal{S}}$, which is computationally expensive and memory-intensive, even $|\mathcal{S}|$ is small; 
\textit{ii}) Directly minimizing the prediction discrepancy across the entire distribution is computationally infeasible due to the curse of dimensionality and the continuous nature of the variables.
%\end{enumerate*}

\subsection{Adversarial Prediction Matching Framework}\label{sec:apm}
To tackle the above challenges, we first posit the existence of a subset $\mathcal{R}_e\subseteq\mathbb{R}^d$ within the feasible domain of $x$, where every element \( x_{e} \in \mathcal{R}_e \) satisfies the following conditions: \textit{i}) it is associated with a high conditional probability \( p(x_{e} | y) \) for its corresponding class \( y \); and \textit{ii}) it induces substantial logits differences between the \( f_{\theta^{\mathcal{T}}} \) and \( f_{\theta_{e}^{\mathcal{S}}} \), where $e\in\{0, \ldots,E-1\}$ refers to the index of student checkpoints collected at each epoch of running $Alg$. We conjecture that such elements can ``dominate" the expectation of logits difference over \( p(x) \), denoted as:
\begin{equation}\label{eq:bound}
	\mathbb{E}_{x_e \in \mathcal{R}_e}[d\left(f_{\theta^\mathcal{T}}(x_e), f_{\theta_e^\mathcal{S}}(x_e)\right)] \geq \mathbb{E}_{x \sim p(x)}[d\left(f_{\theta^\mathcal{T}}(x), f_{\theta_e^\mathcal{S}}(x)\right)]
\end{equation}
%\begin{equation}\label{eq:bound}
%	\mathbb{E}_{x_e \in \mathcal{R}_e}[d\left(f_{\theta^\mathcal{T}}(x_e), f_{\theta_e^\mathcal{S}}(x_e)\right)] \geq \mathbb{E}_{x \in \mathbb{R}^d}[d\left(f_{\theta^\mathcal{T}}(x), f_{\theta_e^\mathcal{S}}(x)\right)]
%\end{equation}

\begin{figure}[t]
	\begin{center}
		{\includegraphics[width=0.95\textwidth]{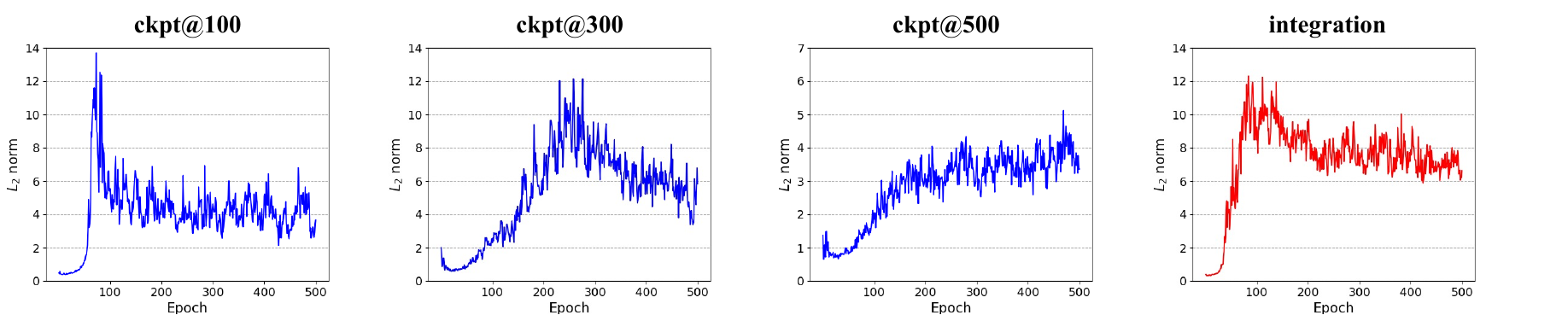}}
		\caption{Trends of $L_2$ norm of $\nabla_\theta\ell(\theta,\mathcal{S})$ with $\mathcal{S}$ derived by Algorithm \ref{alg:framework}, utilizing checkpoints from a specified epoch (left three figures) and assorted epochs (the far right figure) for CIFAR-10, where $|\mathcal{S}|=10\times50$. For example, “ckpt@100" denotes $\mathcal{S}$ is synthesized by referencing student models solely at epoch 100; “integration" utilizes models from epochs 100, 200, 300, 400 and 500.}
		\label{fig:gradnorm}
	\end{center}
\end{figure}

Namely, directly minimizing the expectation of logits difference over $x_e$ can facilitate the prediction alignment of each $\theta_{e}^{\mathcal{S}}$ to the target $\theta^{\mathcal{T}}$ over the original distribution. 
Based on this assumption, we propose to optimize the original objective by a curriculum-based adversarial framework. In each round, we first train a student model from $\theta_0^{\mathcal{S}}$ by the optimizer $Alg$ on $\mathcal{S}=\{(u_i,f_{\theta^{\mathcal{T}}}(u_i))\}$ for $E$ epochs with $d(.,.)$ as the loss function, then alternately force synthetic samples $\{u_i\}_{i=1}^{|\mathcal{S}|}$ to approach ``hard samples"  \( x_{e} \in \mathcal{R}_e, \forall e\in\{0, \ldots,E-1\} \) with the following target:
\begin{equation}\label{eq:target}
	\max_{u} \frac{1}{|\mathcal{S}|} \sum_{i=1}^{|\mathcal{S}|} \sum_{e=0}^{E-1} \log[ p(u_i|y_{u_i}) d\left( f_{\theta^{\mathcal{T}}}\left(u_i\right) , f_{\theta_e^{\mathcal{S}}}\left(u_i\right) \right) ]
\end{equation}
Where, $y_{u_i}$ refers to the class in which $u_i$ is associated for distilling.  However, the inherent complexity in estimating $p(u|y_u)$ due to the high-dimensional nature of data still hinders the optimization of Equation (\ref{eq:target}). To address this issue, we advocate for substituting \( p(u|y_u) \) with \( p(y_u|u) \) for the maximization of the stated target. According to Bayes' theorem, we have \( p(u|y_u) \propto p(y_u|u)p(u) \), presuming a uniform distribution for \( p(y) \). Given the stark contrast between the high-dimensional continuum inherent in \(u\) and the finite discreteness in \(y\), we posit, heuristically, that prioritizing the maximization of \(p(y_u|u)\) is a plausible strategy for the overall maximization of \(p(y_u|u)p(u)\). Regarding the teacher model derived from the original dataset is reliable, we employ \(p(y_u|u; \theta^{\mathcal{T}}) = \text{softmax}(f_{\theta^{\mathcal{T}}}(u))_{y_u}\) as an approximation for \(p(y_u|u)\). Consequently, our loss function designed for updating \(\{u_i\}_{i=1}^{|\mathcal{S}|}\) to maximize the stated target is defined as follows: 
\begin{equation}\label{eq:loss_u}
	\mathcal{L}_u = \frac{1}{|\mathcal{S}|} \sum_{i=1}^{|\mathcal{S}|} \sum_{e=0}^{E-1} -\log\left[d\left(f_{\theta^{\mathcal{T}}}(u_i), f_{\theta_e^{\mathcal{S}}}(u_i)\right)\right] + \alpha H\left(y_{u_i}, p(y_{u_i} | u_i; \theta^{\mathcal{T}})\right)
\end{equation}
Where, \( H\left(y_{u_i}, p(y_{u_i} | u_i; \theta^{\mathcal{T}})\right) = -\sum_{c=1}^{C} y_c \log\left(p(y_c | u_i; \theta^{\mathcal{T}})\right) = -\log\left(p(y_{u_i} | u_i; \theta^{\mathcal{T}})\right) \) denotes the cross-entropy loss of \( \theta^{\mathcal{T}} \) on \( u_i \) w.r.t the associated distilling class \( y_{u_i} \) and $\alpha$ refers to a hyper-parameter for flexible tradeoff. Intuitively, this loss function encourages the synthetic samples to evolve into valuable, hard points, thus contributing beneficially to the overall training process of $\theta^{\mathcal{S}}$.

\textbf{Practical Implementation.\quad}We outline the overall procedure of our proposed Adversarial Prediction Matching (APM) framework in Algorithm \ref{alg:framework}. We initialize \(\mathcal{S}\) by sampling real data from \(\mathcal{T}\), ensuring each \(u_i\) has a high initial probability of \(p(y_{u_i} | u_i; \theta^{\mathcal{T}})\). Notably, to fulfil the inherent generalization requisite, we adopt the strategy from MTT frameworks~\citep{DBLP:conf/cvpr/Cazenavette00EZ22b,DBLP:conf/cvpr/DuJTZ023,DBLP:conf/icml/CuiWSH23}, obtaining a set of teacher models, \(\{\theta^{\mathcal{T}}\}\), prior to commencement. Subsequently, in each iteration, a \(\theta^{\mathcal{T}}\) is sampled from the set to be aligned with a newly initialized \(\theta_0^{\mathcal{S}} \sim p(\theta)\). To constrain the memory overhead, we compute the loss in Equation (\ref{eq:loss_u}) with $K$ select checkpoints, collected at regular intervals during training. We empirically validate in Section \ref{sec:ablation} that this does not compromise the efficacy of our method. 

\textbf{Necessity of Involving Training Snapshots in $\mathcal{L}_u$.\quad}Ideally, a distilled dataset should remain informative throughout the entire model training process, avoiding limitations to specific stages. Nonetheless, our observations indicate that relying on proxy student models from a singular stage can introduce a significant inductive bias. To clarify this, we generate synthetic samples using Algorithm \ref{alg:framework} but with checkpoints from a specified epoch to calculate $\mathcal{L}_u$. Then, the resultant dataset is used to train newly initialized models, and variations in gradient norm relative to model parameters are observed. Figure \ref{fig:gradnorm} reveals a noteworthy trend where peaks in gradient norms are prominent and only appear near the epochs of the selected checkpoints. Conversely, jointly utilizing these checkpoints for synthetic sample generation exhibits a stable pattern throughout training.

%Variations in loss landscape are also discussed in Appendix \ref{app:landscape}.
% ## if anyone ask the bug in algorithm output "=0",it is raised by the package

\begin{algorithm}[t]
	\caption{Dataset distillation via adversarial prediction matching}
	\label{alg:framework}
	\KwIn{An architecture parameterized by $\theta$; A set of well-trained teacher weights $\{\theta^{\mathcal{T}}\}$; Rounds $R$; Epochs $E$; Number of select checkpoints $K$; Model learning rate $\eta$; Synthetic samples learning rate $\gamma$; Mini-batch size $B$}
	\KwOut{Distilled synthetic dataset $\mathcal{S}$}
	Initialize synthetic samples by randomly sampling $\{u_i\}_{i=1}^{|\mathcal{S}|}\subset \{x_i\}_{i=1}^{|\mathcal{T}|}$\;
	\For{$r=1$ \KwTo $R$}{
		Initialize $\theta^{\mathcal{T}}\sim\{\theta^{\mathcal{T}}\}$, $\theta^{\mathcal{S}}_{0}\sim p(\theta)$, and an empty checkpoint list\;
		Randomly sample a mini-batch $\{u^r_i\}_{i=1}^{B}\subseteq \{u_i\}_{i=1}^{|\mathcal{S}|}$\; %and constitute $\mathcal{S}_r=\{(u^r_i, f_{\theta^\mathcal{T}}(u^r_i))\}$\
		\For{$e=1$ \KwTo $E$}{
			Calculate the prediction matching loss: $\mathcal{L}_\theta=\frac{1}{B}\sum_{i=1}^{B}d(f_{\theta^\mathcal{T}}(u^r_i),f_{\theta_{e-1}^\mathcal{S}}(u^r_i))$\;
			Update $\theta^{\mathcal{S}}_{e}=\theta^{\mathcal{S}}_{e-1}-\eta\cdot\nabla_{\theta^{\mathcal{S}}_{e-1}}\mathcal{L}_{\theta}$\;
			\If{$e \mod \lfloor E/K\rfloor = 0$}{
				Store $\theta^{\mathcal{S}}_{e}$ in the checkpoint list\;
			}
		}
		Calculate $\mathcal{L}_u$ in Equation (\ref{eq:loss_u}) with $K$ checkpoints\;
		Update  $u^{r}=u^{r}-\gamma\cdot\nabla_{u^r}\mathcal{L}_{u}$\;
	}
	Calculate $v_i=f_{\theta^{\mathcal{T}}}(u_i), \forall u_i$ and set $\mathcal{S}=\{(u_i,v_i)\}_{i=1}^{|\mathcal{S}|}$\;
\end{algorithm}

\subsection{Superiority Analysis}\label{sec:sup}
\textbf{Low Memory Complexity.\quad}Based on the definition of the loss function proposed for updating synthetic samples in Equation (\ref{eq:loss_u}), all computational operations conducted on each synthetic sample \(u_i\), e.g., the computation of each $d(f_{\theta^{\mathcal{T}}}(u_i), f_{\theta_e^{\mathcal{S}}}(u_i))$ and $H\left(y_{u_i}, p(y_{u_i} | u_i; \theta^{\mathcal{T}})\right)$ (abbreviated as $d^e_i$ and $H_i$, respectively), are integrated by addition. Thanks to this characteristic, two inherent advantages in memory efficiency emerge: \textit{1.} the gradient of $\mathcal{L}_u$ w.r.t each \(u_i\) can be independently derived without interference from other samples of \(u_{j \neq i}\); \textit{2.} the partial derivative of each computation term, e.g.,  \(\frac{\partial d^e_i}{\partial u_i}\) or \(\frac{\partial H_i}{\partial u_i}\), can be independently calculated, and finally updating \(u_i\) based on the accumulated gradients of each term.  Thus, the minimal memory complexity of our method can be expressed as \(\mathcal{O}(\mathcal{G}_{\text{ours}})\), where \(\mathcal{G}_{ours}\) refers to the computation graph size for a single input sample, and it is dominated by \(\frac{\partial d^e_i}{\partial u_i}\) or \(\frac{\partial H_i}{\partial u_i}\) which only involves a trivial backpropagation through models. In contrast, TESLA \citep{DBLP:conf/icml/CuiWSH23} achieves the optimal memory efficiency among MMT frameworks  by reducing the memory complexity of the original \citep{DBLP:conf/cvpr/Cazenavette00EZ22b} from $\mathcal{O}\left(TB\mathcal{G}_{MTT}\right)$ to $\mathcal{O}\left(B\mathcal{G}_{MTT}\right)$, where $T$ and $B$ denote specified updating steps of deriving trajectories and the mini-batch updating size for synthetic data, respectively. Compared with ours, except the mini-batch size $B$, the computation graph size $\mathcal{G}_{MTT}$ is dominated by a complicated Jacobian matrix $\frac{\partial }{\partial u_i}  \nabla_\theta \ell(\theta,\mathcal{S})\in\mathbb{R}^{|\theta|\times d}$. The remarkable memory efficiency of our method enables the distillation of various-size synthetic datasets for large-scale datasets, such as ImageNet-1K, even under memory-constrained circumstances.

\textbf{Customized Tradeoff between Time and Memory.\quad}We show that $\nabla_{u_i} \mathcal{L}_u$ for each $u_i$ can be calculated independently and further decomposed into partial derivatives of each computational term. In practice, this implies multiple model invocations for separate calculations, elongating the runtime. Thus, our algorithm optionally allows computation of the gradients w.r.t multiple $u$ simultaneously, offering a flexible tradeoff between memory and runtime. Further, Section \ref{sec:efficiency} empirically validates our capability to distribute distillation tasks to independent workers (e.g., generating $N$ subsets of size $|\mathcal{S}|/N$ in parallel and aggregating), a feature absent in prior dataset distillation methods.

\textbf{Significantly Reduced Storage vs. MTT.\quad}Our approach only requires matching predictions of the converged teacher, requiring one checkpoint per teacher. However, MTT frameworks necessitate saving checkpoints from every epoch of training a teacher. Therefore, even utilizing the same setup to acquire an equal number of teachers, the storage usage of our method is only a fraction (i.e., 1/epoch) of that of MTT frameworks. Moreover, an ablation study in Section \ref{sec:ablation} demonstrates that the performance of our method obtained with 10 teachers is comparable to that with 100 teachers.

\section{Experiments}
%In this section, we first compare our proposed framework with state-of-the-art solutions of dataset distillation along different distilling protocols. To emphasize the practicability, we illustrate that the synthetic data can be generated distributively and then merged by our proposed framework without a significant performance decline. We then exhibit the superior performance of our generated synthetic data in the cross-architecture generalization ability. After that, we conduct comprehensive ablation studies to learn the influence of hyper-parameters. Finally, we apply our generated synthetic dataset to a practical neural architecture search task to demonstrate its reliability.
\subsection{Experimental Setups}
%\subsubsection{Datasets.}
\textbf{Datasets.\quad}We evaluate the performance of our method on distilling four benchmark datasets, including CIFAR-10/100~\citep{cifar}, Tiny ImageNet~\citep{le2015tiny} and ImageNet-1K~\citep{DBLP:journals/ijcv/RussakovskyDSKS15}. Notably, following the protocol of \citet{zhou2022dataset} and \citet{DBLP:conf/icml/CuiWSH23}, we resize the resolution of ImageNet-1K images to $64\times64$ in our evaluations for fair comparison. We provide further detailed information of benchmark datasets in Appendix \ref{app:datasets}.

\textbf{Evaluation Setup and Baselines.\quad}We employ several recent distillation frameworks as baselines including a short-term gradient matching method Data Siamese Augmentation (DSA) \citep{DBLP:conf/icml/ZhaoB21}, a feature distribution matching method Distribution Matching (DM) \citep{DBLP:journals/corr/abs-2110-04181}, two bi-level-optimization-based methods which are FRePo~\citep{zhou2022dataset} and RCIG~\citep{DBLP:conf/icml/LooHLR23}, and three training trajectory matching methods which are Matching Training Trajectory (MTT)~\citep{DBLP:conf/cvpr/Cazenavette00EZ22b} and its two variations TESLA \citep{DBLP:conf/icml/CuiWSH23} and Flat Trajectory Distillation (FTD) \citep{DBLP:conf/cvpr/DuJTZ023}. 

%Further, HaBa \citep{DBLP:journals/corr/abs-2210-16774} recently proposed a new setup for the dataset distillation, which is known as the dataset factorization that the tested synthetic samples are replaced by the outputs of several trained hallucination networks fed with optimized base samples. Haba shows its superior performances by generally boosting existing frameworks. We also investigate the performance of our method combined with Haba and compare it with MTT+Haba which achieves state-of-the-art performance among multiple benchmark tests.
\textbf{Evaluation Metrics.\quad}To evaluate compared methods, we adhere to the standard protocol prevalent in the existing literature. We train 5 randomly initialized networks from scratch over the distilled dataset and evaluate their performances by reporting the mean and standard deviation of their accuracy on the test set. We use the terminology ``IPC'' (an acronym for ``images per class'') in dataset distillation to indicate the size of distilled datasets. Specifically, we examine an IPC of 50/500/1000 for CIFAR-10 and an IPC of 10/50/500 for CIFAR-100, Tiny ImageNet, and ImageNet-1K.

% And for the dataset factorization, we follow the setup suggested by HaBa that set the number of bases per class ($bpc$) as $ipc$ minus 1 in each configuration when $ipc$ is greater than 1, to 
%offset the storage cost of hallucinator networks.
%\subsubsection{Implementation Details.}
\textbf{Implementation Details.\quad}We implement our method based on the procedure of Algorithm \ref{alg:framework}. Specifically, we set the teacher number of $\{\theta^{\mT}\}$ as 100 to be fairly compared with MTT and its variations TESLA and FTD. We set the distance metric $d$ by $L_1$ distance, the running round $R=2,500$, the number of checkpoints $K=5$ and the coefficient of cross-entropy regularizer $\alpha=0.1$ by default. Further details of our hyper-parameter setup and guidelines are in Appendix \ref{app:hyper}.  
%Manhattan distance is employed as the default distance metric. We also investigate performances achieved by several other distance metrics in Appendix \ref{sec:dist}.
% Follow \citep{DBLP:journals/corr/abs-2210-16774}, we also employ 5 hallucination networks for combing with HaBa. 
% More details about our hyperparameter setup can be found in Appendix \ref{}. 
% And detailed information on combing HaBa with our method can be investigated in Appendix \ref{}. 
For a fair comparison, we stay with all the compared precedents to adopt the 3-layer ConvNet architecture \citep{DBLP:conf/cvpr/GidarisK18} for distilling CIFAR-10 and CIFAR-100, and extend to the 4-layer ConvNet for Tiny ImageNet and ImageNet-1K. Moreover, we also employ ZCA whitening for CIFAR-10 and CIFAR-100 as done by previous works \citep{zhou2022dataset,DBLP:conf/icml/LooHLR23,DBLP:conf/cvpr/Cazenavette00EZ22b,DBLP:conf/cvpr/DuJTZ023,DBLP:conf/icml/CuiWSH23}.
All the experimental results of our method can be obtained on one  RTX4090 GPU, including the extensive ImageNet-1K dataset.

\subsection{Comparison to State-of-the-art}\label{sec:comp-sota}
\textbf{Evaluation on ConvNets.\quad}As illustrated by Table~\ref{tab:comp}, our method demonstrates superior performance than all the baselines among all the evaluation scenarios. Notably, ConvNet models trained on our distilled datasets, which are just $10\%$ the size of the original, manage to achieve, on average, $96\%$ of the test accuracies of models trained on the full CIFAR-10, CIFAR-100, and Tiny ImageNet. Conversely, models trained on equivalent-sized random subsets can only reach $58\%$, while the state-of-the-art trajectory matching framework, FTD, achieves a maximum of $87\%$.
Specifically, the average test accuracies of our method surpass those of FTD by $3.1\%$ on CIFAR-10, $2.1\%$ on CIFAR-100, and $5.8\%$ on Tiny ImageNet. Compared to the state-of-the-art bi-level optimization framework RCIG, our framework not only exhibits superior performances at a smaller IPC but also demonstrates feasibility and effectiveness when generating synthetic sets with a larger IPC. Furthermore, thanks to the memory efficiency of our adversarial prediction matching framework, our method stands as one of the only two capable of successfully distilling the extensive ImageNet-1K dataset, and it additionally exceeds the competitor, TESLA, by $4.4\%$ on average.
A visualization study of our synthetic data is shown in Appendix \ref{app:visual}.

\textbf{Cross-Architecture Generalization.\quad}In Table \ref{tab:cross}, we assess the generalization performance of datasets distilled by our method and by several representative baselines, including DSA, DM, RCIG, and FTD, on unseen model architectures over CIFAR and Tiny ImageNet with an IPC of 50.
We utilize three different neural network architectures for evaluation: VGG11~\citep{DBLP:journals/corr/SimonyanZ14a}, ResNet18~\citep{DBLP:conf/cvpr/HeZRS16} and Vision Transformer (ViT)~\citep{DBLP:conf/iclr/DosovitskiyB0WZ21}. Those models are widely employed for cross-architecture evaluations of recent dataset distillation works~\citep{DBLP:conf/nips/CuiWSH22,DBLP:conf/icml/CuiWSH23,DBLP:conf/cvpr/Cazenavette00EZ23}.
 The results demonstrate that models learned with our synthetic data have superior performances than models trained on datasets synthesized by other baselines. Additionally, compared to the results on ConvNet shown in Table \ref{tab:comp}, we surprisingly noticed larger performance improvements of our method to the state-of-the-art FTD method across multiple unseen models. Moreover, our results obtained on ResNet18 even exceed  the results obtained on ConvNet when testing on CIFAR-100 and Tiny ImageNet.The remarkable transferability demonstrates that our method is more capable of capturing the essential features for learning.
We also report the generalization result of our distilled dataset of ImageNet-1K  in Appendix \ref{app:cag1k}.
\begin{table*}[t]  
	\tiny
	\centering  
	\caption{Comparison of the test accuracy (\%) between ours and other distillation methods on CIFAR10/100, Tiny ImageNet and ImageNet-1K datasets. Entries marked as absent are due to scalability issues. See Appendix \ref{app:issuse} for detailed reasons.}    
	\begin{tabularx}{\linewidth}{>{\centering\arraybackslash}p{1.05cm}>{\centering\arraybackslash}p{0.4cm}|ccccccccc|>{\centering\arraybackslash}p{0.7cm}}  \toprule    Dataset & IPC   & Random & DSA   & DM    & FRePo & RCIG  & MTT   & TESLA & FTD   & \textbf{Ours} & Whole \\  \midrule  \multirow{3}[0]{*}{CIFAR-10} & 50    & 50.2±0.5 & 60.6±0.5 & 63.0±0.4 & 71.7±0.2 & 73.5±0.3 & 71.6±0.2 & 72.6±0.7 & 73.8±0.2 & \textbf{75.0±0.2} & \multirow{3}[0]{*}{85.6±0.2} \\          & 500   & 71.9±0.3 & -     & 74.3±0.2 & -     & -     & 78.6±0.2 & 78.4±0.5     & 78.7±0.2 & \textbf{83.4±0.3} &  \\          & 1000  & 78.2±0.3 & -     & 79.2±0.2 & -     & -     & 79.7±0.2 & 80.3±0.3 & 81.2±0.2 & \textbf{84.6±0.2} &  \\   \midrule  \multirow{3}[0]{*}{CIFAR-100} & 10    & 14.4±0.5 & 32.3±0.3 & 29.7±0.3 & 42.5±0.2 & 44.1±0.4 & 40.1±0.4 & 41.7±0.3 & 43.4±0.3 & \textbf{44.6±0.3} & \multirow{3}[0]{*}{56.8±0.2} \\          & 50    & 30.5±0.3 & 42.8±0.4 & 43.6±0.4 & 44.3±0.2 & 46.7±0.3 & 47.7±0.2 & 47.9±0.3 & 50.7±0.3 & \textbf{53.3±0.2} &  \\          & 100   & 43.1±0.2 & 44.7±0.2 & 47.1±0.4 & -     & 47.4±0.3 & 50.3±0.1 & 51.1±0.3     & 52.8±0.3 & \textbf{55.2±0.2} &  \\   \midrule  \multirow{3}[0]{*}{Tiny ImageNet} & 10    & 4.8±0.3 & 15.9±0.2 & 12.9±0.4  & 25.4±0.2 & 29.4±0.2 & 23.2±0.2 & 14.1±0.3     & 24.5±0.2 & \textbf{30.0±0.3} & \multirow{3}[0]{*}{40.0±0.4} \\          & 50    & 15.1±0.3 & 21.6±0.3 & 21.2±0.3 & -     & -     & 28.0±0.3 & 33.4±0.5     & 31.5±0.3 & \textbf{38.2±0.4} &  \\          & 100   & 24.3±0.3 & -     & 29.4±0.2 & -     & -     & 33.7±0.6 & 34.7±0.2     & 34.5±0.4 & \textbf{39.6±0.2} &  \\   \midrule  \multirow{3}[0]{*}{ImageNet-1K} & 10    & 4.1±0.1 & -     & -     & -     & -     & -     & 17.8±1.3 & -     & \textbf{24.8±0.2} & \multirow{3}[0]{*}{37.2±0.5} \\          & 50    & 16.2±0.8 & -     & -     & -     & -     & -     & 27.9±1.2 & -     & \textbf{30.7±0.3} &  \\          & 100   & 19.5±0.5 & -     & -     & -     & -     & -     & 29.2±1.0     & -     & \textbf{32.6±0.3} &  \\    \bottomrule
	\end{tabularx}%  
	\label{tab:comp}%
\end{table*}%

%We attribute the superior performance of our method to the fact that, unlike previous frameworks, our framework does not aim to make the models trained on the synthetic set accurately predict on a subset of the original data or mimic the model behaviors induced by a subset of the original data, such as gradients. Instead, it treats synthetic data as "hard samples" that effectively assist models in aligning predictions with models trained on original data. 

%\begin{figure}[ht]
%	\begin{center}
%		\centerline
%		{\includegraphics[width=0.48\textwidth]{figs/merged_batches.png}}
%		\caption{The merging effectiveness comparison demonstrated by the training performances of merged synthetic datasets varying with the number of distributively generated batches.}
%		\label{fig:merge}
%	\end{center}
	
%\end{figure}

\subsection{Memory and Time Efficiency Analysis}\label{sec:efficiency}
\begin{figure}[t]
	\centering
	\begin{minipage}{0.43\textwidth}  % Adjust the width of the minipage as needed
		\centering
		\includegraphics[width=\linewidth]{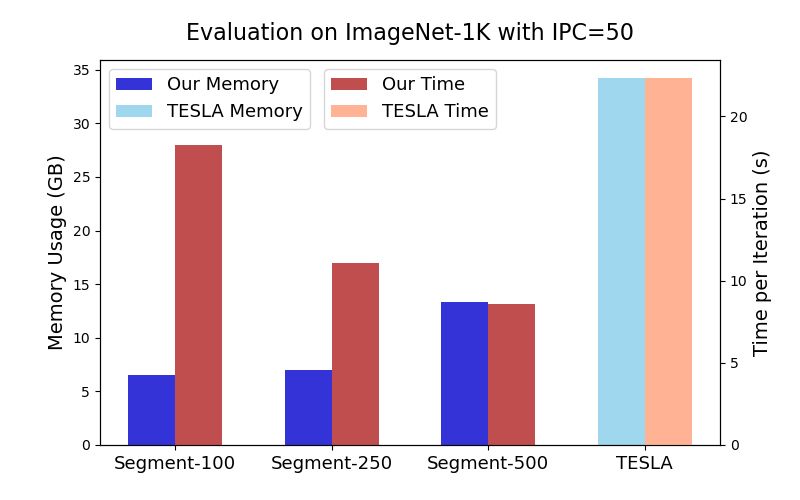} % Adjust the image name and width as needed
		\caption{Flexible tradeoff between memory usage and runtime per iteration of our method compared with TESLA's for distilling ImageNet-1K with an IPC of 50.}
		\label{fig:memory}
	\end{minipage}
	\hspace{0.8cm} % To add space between the two figures
	\begin{minipage}{0.47\textwidth}  % Adjust the width of the minipage as needed
		\centering
		\includegraphics[width=\linewidth]{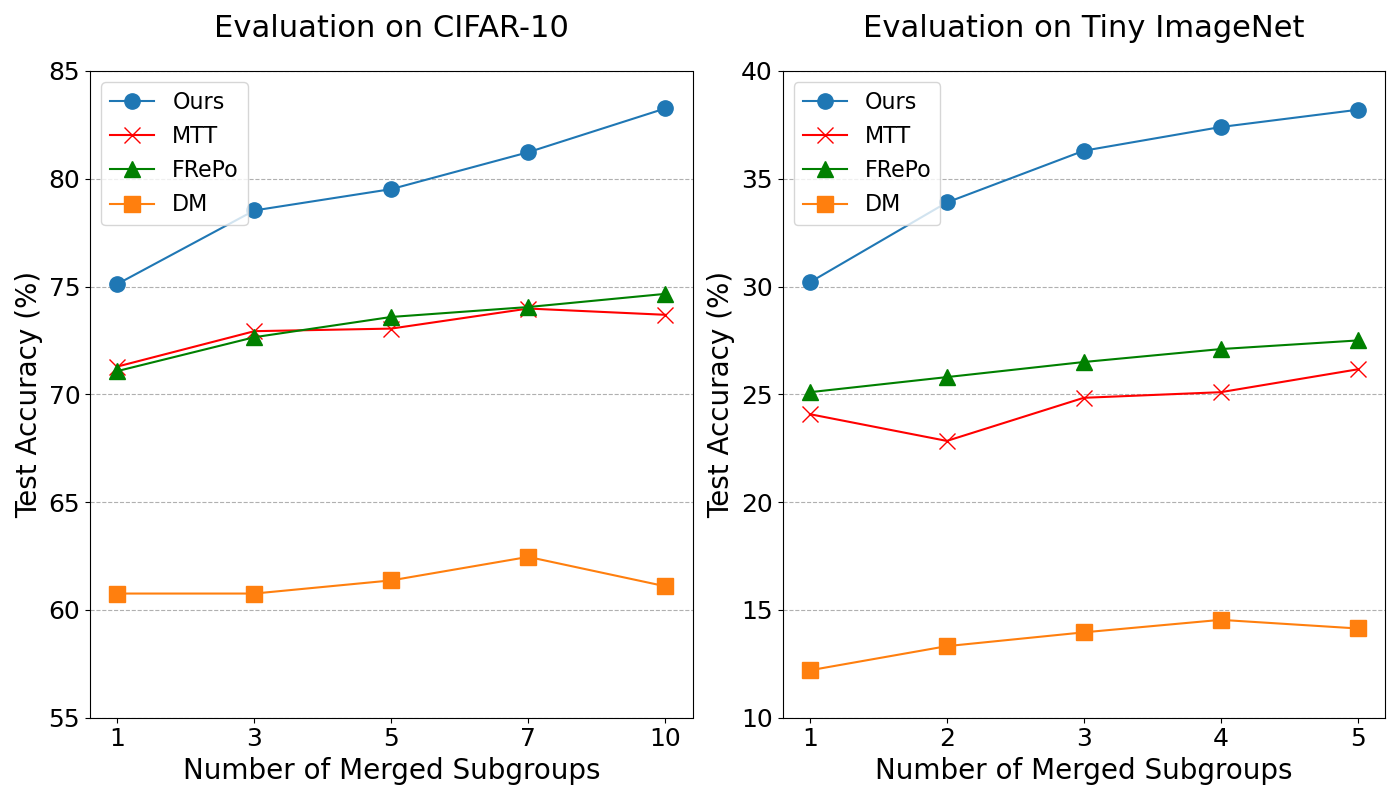} % Adjust the image name and width as needed
		\caption{Test accuracy (\%) varying with merging different numbers of distributively generated sub-batches. IPC of each sub-batch is 50 for CIFAR-10 and 10 for Tiny ImageNet.}
		\label{fig:merge}
	\end{minipage}%
\end{figure}

{\textbf{High Efficiency for Distilling ImageNet-1K.}\quad}To empirically evident the superiorities of the memory complexity and the flexible tradeoff between runtime and memory usage we discussed in Section \ref{sec:sup}, we compare our method with TESLA, which exhibits the optimal memory complexity among trajectory matching frameworks, for distilling ImageNet-1K with an IPC of $50$. We conduct tests on a Tesla A100 GPU (80GB) for a fair comparison.  Figure \ref{fig:memory} represents the averaged peak memory usage and runtime per iteration over 50 rounds collected by running algorithms with the reported default hyper-parameters for obtaining the results shown in Table \ref{tab:comp}. Both methods set the mini-batch size for updating synthetic samples in each iteration as $500$. However, our method allows for an equivalent update by distributing the updates of 500 samples across multiple segments. For instance, “Segment-100” implies dividing 500 samples into five parts and updating sequentially. Through this approach, our method can flexibly tradeoff memory usage and runtime, enabling the distillation of ImageNet-1K with merely 6.5GB of memory and achieving superior distillation performance. Additionally, we observe that even when updating 500 samples at once, our method consumes $2.5\times$  less memory than Tesla. Given that the default running iteration of Tesla is double ours, our overall runtime is only one-fifth of theirs.

{\textbf{Parallelizability of Dataset Distillation.}\quad}To further enhance productivity, a logical approach is to divide the generation of a synthetic dataset and allocate it to multiple workers for parallel execution without compromising effectiveness. We refer to this feature as the parallelizability of dataset distillation methods. We evaluate this feature of our method and several representative frameworks including DM, FRePo and MTT in Figure \ref{fig:merge}. Specifically, we separately craft 10 groups of synthetic data with an IPC of 50 for CIFAR-10 and 5 groups with an IPC of 10 for Tiny ImageNet. We can see from the figure that the performances of models attained on our synthetic datasets persistently increase with the growing number of merged groups. Moreover, we are the only framework that can achieve equivalent performances as the original results shown in Table \ref{tab:comp}. This indicates that, unlike other frameworks, running our method in parallel does not capture repetitive or redundant features.

%  Absent entries of RCIG are attributed to the inability to distil Tiny ImageNet with an IPC of 50.
\begin{table}[t]  
	\tiny
	\centering  
	\caption{Cross-architecture generalization performance in test accuracy (\%) of synthetic datasets distilled for CIFAR-10/100 and Tiny ImageNet at an IPC of 50.}    
	\begin{tabular}{c|c|cccccc}
		%		{c|c|cccccc}     
		\toprule  Dataset   &    Model   & Random & DSA   & DM    & RCIG  & FTD   & \textbf{Ours} \\  \midrule  \multirow{3}[0]{*}{CIFAR-10} & VGG11 & 46.7±0.6 & 51.0±1.1 & 59.2±0.8 & 47.9±0.3 & 59.1±0.2 & \textbf{65.8±1.0} \\          & ResNet18 & 48.1±0.3 & 47.3±1.0 & 57.5±0.9 & 59.5±0.2 & 64.7±0.3 & \textbf{69.2±0.8} \\          & ViT   & 38.9±0.6 & 22.6±1.0 & 27.2±1.6 & 27.7±0.8 & 38.7±0.9 & \textbf{42.2±1.1} \\  \midrule  \multirow{3}[0]{*}{CIFAR-100} & VGG11 & 28.4±0.3 & 29.8±0.1 & 34.6±0.4 & 36.7±0.1 & 42.5±0.2 & \textbf{46.6±0.2} \\          & ResNet18 & 41.0±0.8 & 41.7±0.7 & 37.7±0.5 & 36.5±0.4 & 48.4±0.1 & \textbf{55.2±0.3} \\          & ViT   & 25.5±0.2 & 27.4±0.3 & 28.2±0.6 & 15.7±0.6 & 30.2±0.5 & \textbf{35.2±0.6} \\   \midrule \multirow{3}[0]{*}{Tiny ImageNet} & VGG11 & 21.1±0.3 & 21.5±0.3 & 22.0±0.4 & -     & 27.2±0.3 & \textbf{35.1±0.1} \\          & ResNet18 & 30.7±0.8 & 30.9±1.1 & 20.3±0.4 & -     & 35.7±0.6 & \textbf{40.2±0.2} \\          & ViT   & 17.4±0.6 & 18.6±0.3 & 18.1±0.4 & -     & 21.5±0.1 & \textbf{24.5±0.8} \\   \bottomrule
	\end{tabular}% 
	\label{tab:cross}%
\end{table}%f

\subsection{Ablation Study} \label{sec:ablation}
\begin{figure*}[t]
	\centering
	\subfigure[Teacher number]{
		\label{subfig:expt}
		\begin{minipage}[t]{0.24\linewidth}
			\centering
			\includegraphics[width=1.4in]{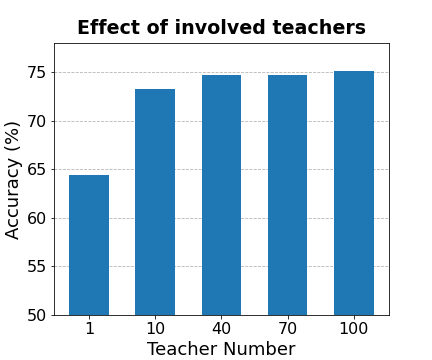}
		\end{minipage}%
	}%
	\subfigure[Checkpoint number]{
		\label{subfig:num}
		\begin{minipage}[t]{0.24\linewidth}
			\centering
			\includegraphics[width=1.4in]{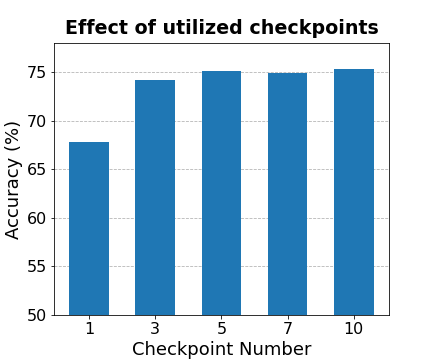}
		\end{minipage}%
	}%
	\subfigure[Mini-batch size]{
		\label{subfig:size}
		\begin{minipage}[t]{0.24\linewidth}
			\centering
			\includegraphics[width=1.4in]{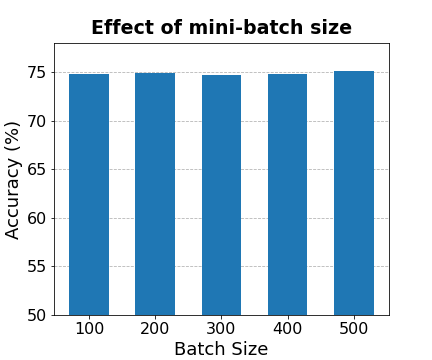}
		\end{minipage}%
	}%
	\subfigure[Updating epoch of $\theta^{\mS}$]{
		\label{subfig:epoch}
		\begin{minipage}[t]{0.24\linewidth}
			\centering
			\includegraphics[width=1.4in]{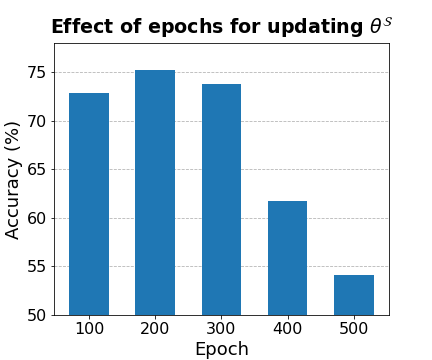}
		\end{minipage}%
	}%
	\centering
	\caption{Ablation studies of our method conducted on CIFAR-10 with an IPC of 50.}
	\vspace{-0.2cm}
	\label{fig:ablation}
\end{figure*}

%In Figure \ref{subfig:expt} \& \ref{subfig:num}, we explore the effectiveness of the two techniques proposed for enhancing the adversarial framework in Sec. \ref{sec:apm} (). We also investigate the stability of model performances varying with the mini-batch size of synthetic data sampled in each iteration (Figure \ref{subfig:size}) and the hyper-parameter tuning process of the epoch number for updating the model weights over the synthetic data (Figure \ref{subfig:epoch}). 
We study the effectiveness or the influence of the following four components of our method. 
\textbf{Number of Pre-trained Teachers:} To enhance the generalization ability, we propose to match prediction with different teachers sampled from a pre-gathered pool $\{\theta^T\}$.  Figure \ref{subfig:expt} reveals a consistent performance increase with the size of $\{\theta^T\}$, emphasizing the importance of dynamically altering the teacher model to circumvent overfitting to a singular model. 
Notably, the result with $10$ teachers is nearly equivalent to that with $100$ teachers, highlighting the potential for computation and storage resource conservation. 
\textbf{Number of Involved Checkpoints:} As illustrated by Figure \ref{subfig:num}, utilizing $K$ checkpoints in Equation (\ref{eq:loss_u}) improves the performance of only using the last checkpoint. Besides, we observe that setting $K$ larger than 5 will not cause significant changes in the performance, allowing equivalent effectiveness with reduced resources.
\textbf{Mini-Batch Size:} In Sec \ref{sec:sup}, we emphasize that the loss function for updating one synthetic sample is computationally independent of others. 
% This property liberated the optimization for a relatively large synthetic dataset when users have strong demand for the performance of the distilled dataset but with limited GPU memories. 
Fig \ref{subfig:size} shows the invariance of performances of synthetic datasets distilled with different mini-batch sizes $B$. Thus, both memory consumption and runtime of each round can be effectively reduced when the mini-batch size is small. 
% \textcolor{blue}{It should be noticed that the tunn of  the hyperparameters of epoch and learning rate for updating the model over the synthetic set had to be redefined to make the model learn adequately.} 
\textbf{Number of Epoch for the Model Update:} We examine the effect of the epoch number $E$ for training student models on the synthetic dataset in Figure \ref{subfig:epoch}. We can see a significant decrease when $E$ is set higher than $300$. This is because a large $E$ will cause most of the $K$ collected checkpoints to be nearly converged. Thus, condensed features derived by using them as proxies lack sufficient knowledge beneficial for the early stages of model training.
Moreover, we comprehensively study the influence of varying the coefficient of cross-entropy regularizer, $\alpha$, over the four benchmark datasets in Appendix \ref{app:alpha}.

\begin{table}[t] 
	\centering
	\small
	\caption{Comparison of Spearman's ranking correlation of NAS on CIFAR-10.  Results are obtained from models trained on various distilled datasets with an IPC of 50 (i.e., 1\% of the original dataset) over 200 epochs. For reference, the correlation achieved by training models on the full original dataset for 20 epochs stands at 0.52, albeit requiring $10\times$ the search time than ours.}    
	\begin{tabular}{cccccc}
		\toprule    Random & DSA   & DM    & MTT   & FRePo  & \textbf{Ours} \\   \midrule 0.18±0.06 & 0.12±0.07 & 0.13±0.06 & 0.31±0.01 & -0.07±0.07 & \textbf{0.47±0.03} \\    \bottomrule
	\end{tabular}%  
\label{tab:nas}%
\end{table}%

\subsection{Application: Neural Architecture Search}

To explore the practicality of our method, we utilize the distilled datasets as a proxy for the intricate Neural Architecture Search (NAS) task. For this, we adopt the same NAS-Bench-201 task \citep{DBLP:conf/iclr/Dong020}, previously employed by a recent dataset distillation benchmark \citep{DBLP:conf/nips/CuiWSH22}.
Specifically, we randomly sample 100 networks from NAS-Bench-201, in which the search space is comprised of 15,625 networks with ground-truth performances collected by training on the entire training dataset of CIFAR-10 for 50 epochs under 5 random seeds, and ranked according to their average accuracy on a held-out validation set of 10k images. We compare our methods with various baselines including Random (a randomly selected subset), DSA, DM, MTT and FRePo by training the 100 different architectural models over synthetic datasets with an IPC of 50 for 200 epochs. The Spearman’s correlation between the rankings derived from the synthetic datasets and the ground-truth ranking is used as the evaluation metric.
Comparison results are reported in Table \ref{tab:nas}. Notably, due to the inherent variability in training, an average ranking correlation of 0.76 is achieved locally even when adhering to the ground-truth protocol. We can clearly see from the table that the ranking correlation attained by our method markedly surpasses other baselines. Especially, although MTT and FRePo's test accuracies in Table \ref{tab:comp} are on par with ours for an IPC of 50 on CIFAR-10, the efficiency of our method for this NAS task is evident. Moreover, the correlation achieved by training models on the entire original training dataset of CIFAR-10 for 20 epochs is 0.52. However, our distilled datasets deliver comparable performance in just one-tenth of the search time. 
%These observations validate the potential of our method in enhancing the efficiency of NAS tasks.
\section{Conclusions}
In this work, we introduce an innovative formulation for dataset distillation via prediction matching, realized with an adversarial framework, devised for effective problem resolution. Comprehensive experiments validate that our distilled datasets surpass contemporary benchmarks, showcasing exceptional cross-architecture capabilities. Furthermore, we illustrate the superior memory efficiency and practical viability of large-scale dataset distillation of our method.
For future research, we intend to explore advancements in efficiency by diminishing its reliance on the number of teacher models.

%\subsubsection*{Author Contributions}
%If you'd like to, you may include  a section for author contributions as is done
%in many journals. This is optional and at the discretion of the authors.
%
%\subsubsection*{Acknowledgments}
%Use unnumbered third level headings for the acknowledgments. All
%acknowledgments, including those to funding agencies, go at the end of the paper.

\bibliographystyle{iclr2024_conference}
\bibliography{ref}

\begin{thebibliography}{38}
\providecommand{\natexlab}[1]{#1}
\providecommand{\url}[1]{\texttt{#1}}
\expandafter\ifx\csname urlstyle\endcsname\relax
  \providecommand{\doi}[1]{doi: #1}\else
  \providecommand{\doi}{doi: \begingroup \urlstyle{rm}\Url}\fi

\bibitem[Aljundi et~al.(2019{\natexlab{a}})Aljundi, Belilovsky, Tuytelaars,
  Charlin, Caccia, Lin, and Page{-}Caccia]{DBLP:conf/nips/AljundiBTCCLP19}
Rahaf Aljundi, Eugene Belilovsky, Tinne Tuytelaars, Laurent Charlin, Massimo
  Caccia, Min Lin, and Lucas Page{-}Caccia.
\newblock Online continual learning with maximal interfered retrieval.
\newblock In \emph{NeurIPS}, pp.\  11849--11860, 2019{\natexlab{a}}.

\bibitem[Aljundi et~al.(2019{\natexlab{b}})Aljundi, Lin, Goujaud, and
  Bengio]{DBLP:conf/nips/AljundiLGB19}
Rahaf Aljundi, Min Lin, Baptiste Goujaud, and Yoshua Bengio.
\newblock Gradient based sample selection for online continual learning.
\newblock In \emph{NeurIPS}, pp.\  11816--11825, 2019{\natexlab{b}}.

\bibitem[Baik et~al.(2020)Baik, Choi, Choi, Kim, and
  Lee]{DBLP:conf/nips/BaikCCKL20}
Sungyong Baik, Myungsub Choi, Janghoon Choi, Heewon Kim, and Kyoung~Mu Lee.
\newblock Meta-learning with adaptive hyperparameters.
\newblock In \emph{NeurIPS}, 2020.

\bibitem[Bergstra et~al.(2011)Bergstra, Bardenet, Bengio, and
  K{\'{e}}gl]{DBLP:conf/nips/BergstraBBK11}
James Bergstra, R{\'{e}}mi Bardenet, Yoshua Bengio, and Bal{\'{a}}zs
  K{\'{e}}gl.
\newblock Algorithms for hyper-parameter optimization.
\newblock In \emph{{NIPS}}, pp.\  2546--2554, 2011.

\bibitem[Bohdal et~al.(2020)Bohdal, Yang, and
  Hospedales]{DBLP:journals/corr/abs-2006-08572}
Ondrej Bohdal, Yongxin Yang, and Timothy~M. Hospedales.
\newblock Flexible dataset distillation: Learn labels instead of images.
\newblock \emph{CoRR}, abs/2006.08572, 2020.

\bibitem[Cazenavette et~al.(2022)Cazenavette, Wang, Torralba, Efros, and
  Zhu]{DBLP:conf/cvpr/Cazenavette00EZ22b}
George Cazenavette, Tongzhou Wang, Antonio Torralba, Alexei~A. Efros, and
  Jun{-}Yan Zhu.
\newblock Dataset distillation by matching training trajectories.
\newblock In \emph{{CVPR}}, pp.\  10708--10717. {IEEE}, 2022.

\bibitem[Cazenavette et~al.(2023)Cazenavette, Wang, Torralba, Efros, and
  Zhu]{DBLP:conf/cvpr/Cazenavette00EZ23}
George Cazenavette, Tongzhou Wang, Antonio Torralba, Alexei~A. Efros, and
  Jun{-}Yan Zhu.
\newblock Generalizing dataset distillation via deep generative prior.
\newblock In \emph{{CVPR}}, pp.\  3739--3748. {IEEE}, 2023.

\bibitem[Chen et~al.(2010)Chen, Welling, and Smola]{DBLP:conf/uai/ChenWS10}
Yutian Chen, Max Welling, and Alexander~J. Smola.
\newblock Super-samples from kernel herding.
\newblock In \emph{{UAI}}, pp.\  109--116. {AUAI} Press, 2010.

\bibitem[Cui et~al.(2022)Cui, Wang, Si, and Hsieh]{DBLP:conf/nips/CuiWSH22}
Justin Cui, Ruochen Wang, Si~Si, and Cho{-}Jui Hsieh.
\newblock {DC-BENCH:} dataset condensation benchmark.
\newblock In \emph{NeurIPS}, 2022.

\bibitem[Cui et~al.(2023)Cui, Wang, Si, and Hsieh]{DBLP:conf/icml/CuiWSH23}
Justin Cui, Ruochen Wang, Si~Si, and Cho{-}Jui Hsieh.
\newblock Scaling up dataset distillation to imagenet-1k with constant memory.
\newblock In \emph{{ICML}}, volume 202 of \emph{Proceedings of Machine Learning
  Research}, pp.\  6565--6590. {PMLR}, 2023.

\bibitem[Dong \& Yang(2020)Dong and Yang]{DBLP:conf/iclr/Dong020}
Xuanyi Dong and Yi~Yang.
\newblock Nas-bench-201: Extending the scope of reproducible neural
  architecture search.
\newblock In \emph{{ICLR}}. OpenReview.net, 2020.

\bibitem[Dosovitskiy et~al.(2021)Dosovitskiy, Beyer, Kolesnikov, Weissenborn,
  Zhai, Unterthiner, Dehghani, Minderer, Heigold, Gelly, Uszkoreit, and
  Houlsby]{DBLP:conf/iclr/DosovitskiyB0WZ21}
Alexey Dosovitskiy, Lucas Beyer, Alexander Kolesnikov, Dirk Weissenborn,
  Xiaohua Zhai, Thomas Unterthiner, Mostafa Dehghani, Matthias Minderer, Georg
  Heigold, Sylvain Gelly, Jakob Uszkoreit, and Neil Houlsby.
\newblock An image is worth 16x16 words: Transformers for image recognition at
  scale.
\newblock In \emph{{ICLR}}. OpenReview.net, 2021.

\bibitem[Du et~al.(2023)Du, Jiang, Tan, Zhou, and Li]{DBLP:conf/cvpr/DuJTZ023}
Jiawei Du, Yidi Jiang, Vincent Y.~F. Tan, Joey~Tianyi Zhou, and Haizhou Li.
\newblock Minimizing the accumulated trajectory error to improve dataset
  distillation.
\newblock In \emph{{CVPR}}, pp.\  3749--3758. {IEEE}, 2023.

\bibitem[Fang et~al.(2019)Fang, Song, Shen, Wang, Chen, and
  Song]{DBLP:journals/corr/abs-1912-11006}
Gongfan Fang, Jie Song, Chengchao Shen, Xinchao Wang, Da~Chen, and Mingli Song.
\newblock Data-free adversarial distillation.
\newblock \emph{CoRR}, abs/1912.11006, 2019.

\bibitem[Fu et~al.(2020)Fu, Geng, Duan, Zhuang, Yuan, Trischler, Lin, Pal, and
  Dong]{role-kd}
Jie Fu, Xue Geng, Zhijian Duan, Bohan Zhuang, Xingdi Yuan, Adam Trischler, Jie
  Lin, Chris Pal, and Hao Dong.
\newblock Role-wise data augmentation for knowledge distillation.
\newblock 2020.

\bibitem[Gidaris \& Komodakis(2018)Gidaris and
  Komodakis]{DBLP:conf/cvpr/GidarisK18}
Spyros Gidaris and Nikos Komodakis.
\newblock Dynamic few-shot visual learning without forgetting.
\newblock In \emph{{CVPR}}, pp.\  4367--4375. Computer Vision Foundation /
  {IEEE} Computer Society, 2018.

\bibitem[He et~al.(2016)He, Zhang, Ren, and Sun]{DBLP:conf/cvpr/HeZRS16}
Kaiming He, Xiangyu Zhang, Shaoqing Ren, and Jian Sun.
\newblock Deep residual learning for image recognition.
\newblock In \emph{{CVPR}}, pp.\  770--778. {IEEE} Computer Society, 2016.

\bibitem[Hinton et~al.(2015)Hinton, Vinyals, and
  Dean]{DBLP:journals/corr/HintonVD15}
Geoffrey~E. Hinton, Oriol Vinyals, and Jeffrey Dean.
\newblock Distilling the knowledge in a neural network.
\newblock \emph{CoRR}, abs/1503.02531, 2015.

\bibitem[Jin et~al.(2019)Jin, Song, and Hu]{DBLP:conf/kdd/JinSH19}
Haifeng Jin, Qingquan Song, and Xia Hu.
\newblock Auto-keras: An efficient neural architecture search system.
\newblock In \emph{{KDD}}, pp.\  1946--1956. {ACM}, 2019.

\bibitem[Kim et~al.(2022)Kim, Kim, Oh, Yun, Song, Jeong, Ha, and
  Song]{DBLP:conf/icml/KimKOYSJ0S22}
Jang{-}Hyun Kim, Jinuk Kim, Seong~Joon Oh, Sangdoo Yun, Hwanjun Song, Joonhyun
  Jeong, Jung{-}Woo Ha, and Hyun~Oh Song.
\newblock Dataset condensation via efficient synthetic-data parameterization.
\newblock In \emph{{ICML}}, volume 162 of \emph{Proceedings of Machine Learning
  Research}, pp.\  11102--11118. {PMLR}, 2022.

\bibitem[Krizhevsky \& Hinton(2009)Krizhevsky and Hinton]{cifar}
Alex Krizhevsky and Geoffrey Hinton.
\newblock Learning multiple layers of features from tiny images.
\newblock \url{https://www.cs.toronto.edu/~kriz/cifar.html}, 2009.
\newblock Accessed: March 1, 2023.

\bibitem[Le \& Yang(2015)Le and Yang]{le2015tiny}
Ya~Le and Xuan Yang.
\newblock Tiny imagenet visual recognition challenge.
\newblock \emph{CS 231N}, 7\penalty0 (7):\penalty0 3, 2015.

\bibitem[Loo et~al.(2023)Loo, Hasani, Lechner, and
  Rus]{DBLP:conf/icml/LooHLR23}
Noel Loo, Ramin~M. Hasani, Mathias Lechner, and Daniela Rus.
\newblock Dataset distillation with convexified implicit gradients.
\newblock In \emph{{ICML}}, volume 202 of \emph{Proceedings of Machine Learning
  Research}, pp.\  22649--22674. {PMLR}, 2023.

\bibitem[Nguyen et~al.(2021)Nguyen, Chen, and Lee]{DBLP:conf/iclr/NguyenCL21}
Timothy Nguyen, Zhourong Chen, and Jaehoon Lee.
\newblock Dataset meta-learning from kernel ridge-regression.
\newblock In \emph{{ICLR}}. OpenReview.net, 2021.

\bibitem[Pham et~al.(2018)Pham, Guan, Zoph, Le, and
  Dean]{DBLP:journals/corr/abs-1802-03268}
Hieu Pham, Melody~Y. Guan, Barret Zoph, Quoc~V. Le, and Jeff Dean.
\newblock Efficient neural architecture search via parameter sharing.
\newblock \emph{CoRR}, abs/1802.03268, 2018.

\bibitem[Rebuffi et~al.(2017)Rebuffi, Kolesnikov, Sperl, and
  Lampert]{DBLP:conf/cvpr/RebuffiKSL17}
Sylvestre{-}Alvise Rebuffi, Alexander Kolesnikov, Georg Sperl, and Christoph~H.
  Lampert.
\newblock icarl: Incremental classifier and representation learning.
\newblock In \emph{{CVPR}}, pp.\  5533--5542. {IEEE} Computer Society, 2017.

\bibitem[Russakovsky et~al.(2015)Russakovsky, Deng, Su, Krause, Satheesh, Ma,
  Huang, Karpathy, Khosla, Bernstein, Berg, and
  Fei{-}Fei]{DBLP:journals/ijcv/RussakovskyDSKS15}
Olga Russakovsky, Jia Deng, Hao Su, Jonathan Krause, Sanjeev Satheesh, Sean Ma,
  Zhiheng Huang, Andrej Karpathy, Aditya Khosla, Michael~S. Bernstein,
  Alexander~C. Berg, and Li~Fei{-}Fei.
\newblock Imagenet large scale visual recognition challenge.
\newblock \emph{Int. J. Comput. Vis.}, 115\penalty0 (3):\penalty0 211--252,
  2015.

\bibitem[Simonyan \& Zisserman(2015)Simonyan and
  Zisserman]{DBLP:journals/corr/SimonyanZ14a}
Karen Simonyan and Andrew Zisserman.
\newblock Very deep convolutional networks for large-scale image recognition.
\newblock In \emph{{ICLR}}, 2015.

\bibitem[Toneva et~al.(2019)Toneva, Sordoni, des Combes, Trischler, Bengio, and
  Gordon]{DBLP:conf/iclr/TonevaSCTBG19}
Mariya Toneva, Alessandro Sordoni, Remi~Tachet des Combes, Adam Trischler,
  Yoshua Bengio, and Geoffrey~J. Gordon.
\newblock An empirical study of example forgetting during deep neural network
  learning.
\newblock In \emph{{ICLR} (Poster)}. OpenReview.net, 2019.

\bibitem[Wang et~al.(2022)Wang, Zhao, Peng, Zhu, Yang, Wang, Huang, Bilen,
  Wang, and You]{DBLP:conf/cvpr/WangZPZYWHBWY22}
Kai Wang, Bo~Zhao, Xiangyu Peng, Zheng Zhu, Shuo Yang, Shuo Wang, Guan Huang,
  Hakan Bilen, Xinchao Wang, and Yang You.
\newblock {CAFE:} learning to condense dataset by aligning features.
\newblock In \emph{{CVPR}}, pp.\  12186--12195. {IEEE}, 2022.

\bibitem[Wang et~al.(2018)Wang, Zhu, Torralba, and
  Efros]{DBLP:journals/corr/abs-1811-10959}
Tongzhou Wang, Jun{-}Yan Zhu, Antonio Torralba, and Alexei~A. Efros.
\newblock Dataset distillation.
\newblock \emph{CoRR}, abs/1811.10959, 2018.

\bibitem[Yin et~al.(2020)Yin, Molchanov, Alvarez, Li, Mallya, Hoiem, Jha, and
  Kautz]{yin2020dreaming}
Hongxu Yin, Pavlo Molchanov, Jose~M. Alvarez, Zhizhong Li, Arun Mallya, Derek
  Hoiem, Niraj~K Jha, and Jan Kautz.
\newblock Dreaming to distill: Data-free knowledge transfer via deepinversion.
\newblock In \emph{The IEEE/CVF Conf. Computer Vision and Pattern Recognition
  (CVPR)}, 2020.

\bibitem[Yin et~al.(2023)Yin, Xing, and
  Shen]{DBLP:journals/corr/abs-2306-13092}
Zeyuan Yin, Eric~P. Xing, and Zhiqiang Shen.
\newblock Squeeze, recover and relabel: Dataset condensation at imagenet scale
  from {A} new perspective.
\newblock \emph{CoRR}, abs/2306.13092, 2023.

\bibitem[Yuan et~al.(2020)Yuan, Tay, Li, Wang, and Feng]{yuan2020revisiting}
Li~Yuan, Francis~EH Tay, Guilin Li, Tao Wang, and Jiashi Feng.
\newblock Revisiting knowledge distillation via label smoothing regularization.
\newblock In \emph{Proceedings of the IEEE/CVF Conference on Computer Vision
  and Pattern Recognition}, pp.\  3903--3911, 2020.

\bibitem[Zhao \& Bilen(2021{\natexlab{a}})Zhao and
  Bilen]{DBLP:conf/icml/ZhaoB21}
Bo~Zhao and Hakan Bilen.
\newblock Dataset condensation with differentiable siamese augmentation.
\newblock In \emph{{ICML}}, volume 139 of \emph{Proceedings of Machine Learning
  Research}, pp.\  12674--12685. {PMLR}, 2021{\natexlab{a}}.

\bibitem[Zhao \& Bilen(2021{\natexlab{b}})Zhao and
  Bilen]{DBLP:journals/corr/abs-2110-04181}
Bo~Zhao and Hakan Bilen.
\newblock Dataset condensation with distribution matching.
\newblock \emph{CoRR}, abs/2110.04181, 2021{\natexlab{b}}.

\bibitem[Zhao et~al.(2021)Zhao, Mopuri, and Bilen]{DBLP:conf/iclr/ZhaoMB21}
Bo~Zhao, Konda~Reddy Mopuri, and Hakan Bilen.
\newblock Dataset condensation with gradient matching.
\newblock In \emph{{ICLR}}. OpenReview.net, 2021.

\bibitem[Zhou et~al.(2022)Zhou, Nezhadarya, and Ba]{zhou2022dataset}
Yongchao Zhou, Ehsan Nezhadarya, and Jimmy Ba.
\newblock Dataset distillation using neural feature regression.
\newblock In \emph{NeurIPS}, 2022.

\end{thebibliography}

\appendix
\newpage
\renewcommand{\thetable}{A\arabic{table}}
\section{Appendix}
%\subsection{Limitations and Future Work}

%\subsection{Loss Landscape of Involving Proxy Students at Different Stage}\label{app:landscape}
%a
\subsection{Benchmark Datasets}\label{app:datasets}
We provide further details of four employed benchmark datasets  including CIFAR-10/100~\citep{cifar}, Tiny ImageNet~\citep{le2015tiny} and ImageNet-1K~\citep{DBLP:journals/ijcv/RussakovskyDSKS15}. For the low-resolution datasets CIFAR-10 and CIFAR-100, there are $5000\times10$ and $500\times100$ training samples, and $1000\times10$ and $100\times100$ testing samples that are under $32\times32$ resolution in 3-channels RGB format, respectively. For high-resolution datasets, Tiny ImageNet comprises $500\times200$ training samples and $50\times200$ testing samples under $64\times64$ resolution with 3 RGB channels in total. Besides, ImageNet-1K is a widely employed large-scale dataset that includes 1,000 classes, encompassing 1,281,167 training images and 50,000 testing images with 3 RGB channels. Following the protocol of \citet{zhou2022dataset} and \citet{DBLP:conf/icml/CuiWSH23}, we resize the resolution of ImageNet-1K images to $64\times64$.

\subsection{Scalability Issues of Competitors}\label{app:issuse}
In this section, we clarify the reasons behind the absent entries for various baseline methods in Table \ref{tab:comp}. For DSA, achieving results for distillation tasks with an IPC exceeding 50 is unattainable due to the lack of official implementation support in such scenarios, where each iteration's runtime becomes prohibitively long (e.g., exceeding 5 minutes). Distilling ImageNet-1K with an IPC of 10 induces an Out-Of-Memory (OOM) error, even on a Tesla A100 GPU (80GB).
The absence of DM entries for distilling ImageNet-1K is also attributed to OOM errors, resulting from the extensive number of classes. For FRePo, there are two constraining issues: first, its official implementation does not support an IPC above 50; second, the computation of the Gram matrix of features from proxy models on real and synthetic samples demands significant memory, constraining its application on large-scale datasets like Tiny ImageNet or ImageNet-1K.
Concerning RCIG, it manages to evade OOM issues by minimizing the mini-batch size for updating synthetic samples, but our empirical results indicate that this compromises the method's efficacy. The model trained with such a distilled dataset yields inferior performance compared to even smaller synthetic datasets.
Lastly, for MTT and FTD, the memory consumption escalates notably with the stipulated updating steps for aligning trajectories of teacher models, making the acquisition of reasonable results challenging with limited memory resources.

\subsection{Cross-Architecture Generalization on ImageNet-1K}\label{app:cag1k}
We examine and contrast the cross-architecture generalization capabilities of our synthetic datasets, distilled for ImageNet-1K with an IPC of 50, against the sole valid competitor, TESLA \citep{DBLP:conf/icml/CuiWSH23}, as detailed in Table \ref{tab:cross1k}. The outcomes indicate that our approach significantly elevates the performance on randomly sampled sub-batches of equivalent size from the original dataset. Notably, the achievements on VGG11 and ResNet18 even exceed those obtained on the proxy architecture (i.e., ConvNet), evidencing reliable applicability across various unseen convolutional models. The modest advancement on ViT can likely be ascribed to the disparities in receptive fields between ViT's segmented patches and ConvNet’s convolutional windows. Addressing this deficiency is considered a pivotal avenue for our future endeavours.

\begin{table}[htbp] 
	\small
	 \centering  
	 \caption{Cross-architecture generalization performance in test accuracy (\%) of synthetic datasets distilled for ImageNet-1K at an IPC of 50.}    
	 \begin{tabular}{c|cccc}    \toprule      &       & Random & TESLA & \textbf{Ours} \\   \midrule \multirow{4}[0]{*}{ImageNet-1K} & ConvNetD4 & 16.2±0.8 & 26.7±1.2 & \textbf{30.7±0.3} \\          & VGG11 & 15.3±0.2 & 24.6±0.8 & \textbf{31.1±0.1} \\          & ResNet18 & 18.7±1.2 & 28.7±0.5 & \textbf{33.2±0.5} \\          & ViT   & 9.5±0.3 & 14.3±0.5 & \textbf{11.7±1.2} \\  \bottomrule  \end{tabular}%  
	 \label{tab:cross1k}%
 \end{table}%

\subsection{Comparison of Generalization Among Different Proxy Model Architectures}\label{app:proxy}
In our formal implementation, we stay with precedents~\cite{DBLP:conf/cvpr/Cazenavette00EZ22b,DBLP:conf/icml/ZhaoB21,DBLP:conf/iclr/ZhaoMB21,DBLP:conf/cvpr/WangZPZYWHBWY22} and adopt the commonly used ConvNet architecture as the proxy model for both teacher models and student models in the dataset distillation tasks. Here, we also investigate the feasibility of using some other CNN-based architectures as the proxy model to generate distilled datasets and compare their inter-model generalization. We employ another two CNN-based network architectures for comparison including VGG11~\citep{DBLP:journals/corr/SimonyanZ14a} and ResNet18~\citep{DBLP:conf/cvpr/HeZRS16} and compare the generalization of the distilled dataset generated respectively on different architectures on CIFAR-10 and Tiny ImageNet with an IPC of 50. For reference, the test accuracies (\%) of ConvNet, VGG11 and ResNet18 training on the whole original datasets are $85.6\pm0.2$, $88.6\pm0.6$ and $93.4\pm0.5$ over CIFAR-10, and are $40.0\pm0.4$, $46.8\pm0.6$ and $62.4\pm1.2$ over Tiny ImageNet, respectively.  
In Table \ref{tab:proxy}, the distilled datasets are generated with one architecture (denoted as ``proxy'') and then evaluated on another architecture (denoted as ``evaluation'') by training a model from scratch and testing on real testing data.  The findings from these experiments unveil several intriguing, and at times, unexpected insights. First, a model trained using data derived from ResNet18 and evaluated on the same architecture, remarkably achieves a test accuracy of $54.8\pm0.6$, considerably surpassing results obtained from datasets generated by the other two models. \textbf{Importantly, this outcome also significantly surpasses the result (i.e., $41.1\pm0.4$) reported in \citet{DBLP:journals/corr/abs-2306-13092}}, a recent study dedicated to generating distilled datasets specifically for models equipped with batch normalization (e.g., ResNets) on Tiny ImageNet and ImageNet-1K. However, second, we also identified relatively poorer generalization performances of these synthetic datasets on other models, a phenomenon that aligns with the observations made in \citet{DBLP:journals/corr/abs-2306-13092}.
Third, it is observed that models of relative complexity, such as VGG11 and ResNet18, demonstrate superior generalization on Tiny ImageNet as compared to their performance on CIFAR-10. We hypothesize this may attributed to the potential overfitting of complex models on simpler datasets, leading to the injection of obscure patterns during the synthesis of samples via backpropagation. The information contained by patterns may not be recognizable by other models. Similar findings have been noted in preceding studies on dataset distillation~\citep{DBLP:conf/iclr/ZhaoMB21,DBLP:conf/icml/ZhaoB21,DBLP:journals/corr/abs-2110-04181}. Such observations prompt us to delve deeper into the underlying causes of these phenomena and to explore methodologies to more effectively integrate sophisticated models with advanced predictive capabilities in dataset distillation processes. We consider the exploration of these notions as pivotal focal points for our impending research endeavours in the dataset distillation.

%% Table generated by Excel2LaTeX from sheet 'Sheet1'
%\begin{table}[htbp]  
%	\centering  
%	\small
%	\setlength{\tabcolsep}{2.5mm}
%	\caption{Comparison of cross-architecture generalization performances (\%) on CIFAR-10 with $ipc=50$. The distilled datasets are learned on one architecture (C) and then tested on another architecture (T).}    
%\begin{tabular}{c|cccc}     \toprule   C \textbackslash T    & ConvNet & AlexNet & VGG11 & ResNet18 \\ \midrule    ConvNet & 74.5±0.2  & 63.2±1.3  & 64.5±0.8  & 67.2±0.5  \\    AlexNet & 61.0±0.2     & 68.7±1.1  & 61.1±0.1    & 63.8±0.7 \\    VGG11 & 55.2±0.3     & 54.1±0.8     & 58.2±0.5     & 54.3±0.4 \\    ResNet18 & 61.7±0.3     & 59.1±0.3     & 59.7±0.8     & 63.2±0.9 \\    \bottomrule
%\end{tabular}% 
%\label{tab:proxy}%
%\end{table}%

\begin{table}[htbp] 
	\small
	\centering 
	\caption{Comparison of cross-architecture generalization performances (\%) of synthetic datasets generated by different proxy architectures for CIFAR-10 and Tiny ImageNet with an IPC of 50.}    
	\begin{tabular}{c|ccc|ccc}    \toprule      & \multicolumn{3}{c|}{CIFAR-10} & \multicolumn{3}{c}{Tiny ImageNet} \\  \midrule  Proxy\textbackslash{}Evaluation & ConvNet & VGG11 & ResNet18 & ConvNet & VGG11 & ResNet18 \\ \midrule  ConvNet & 75.0±0.2 & 65.8±1.0 & 69.2±0.8 & 30.0±0.3 & 35.1±0.1 & 40.2±0.2 \\    VGG11 & 55.2±0.3 & 54.1±0.8 & 54.3±0.4 & 32.8±0.1 & 36.6±0.6 & 41.8±0.4 \\    ResNet18 & 61.7±0.3 & 59.7±0.8 & 63.2±0.9 & 26.9±0.3 & 32.3±0.4 & 54.8±0.6 \\   \bottomrule \end{tabular}% 
	\label{tab:proxy}%
\end{table}%

\subsection{Ablation Study on the Coefficient of Cross-Entropy Regularizer}\label{app:alpha}
We carry out an exhaustive ablation study concerning the coefficient of the cross-entropy regularizer, $\alpha$, as defined in Equation (\ref{eq:loss_u}). In Table \ref{tab:ce}, we examine the performance of ConvNet models in response to variations in $\alpha$ across four benchmark datasets: CIFAR-10 with an IPC of 50 and CIFAR-100, Tiny ImageNet, and ImageNet-1K, each with an IPC of 10. Our findings reveal that increasing $\alpha$ above $0$ yields modest improvements, attesting to the efficacy of constraining synthetic samples to maintain a high probability of $p(x|y;\theta^T)$. Moreover, the performance appears to remain robust against alterations in $\alpha$, indicating the stability of this parameter. This stability proves beneficial during the hyperparameter tuning phase, especially when adapting our approach to diverse distillation tasks.

\begin{table}[htbp]  
	\small
	\centering  
	\caption{Ablation study on the coefficient of cross-entropy regularizer $\alpha$ over the four benchmark datasets with an IPC of 50 for CIFAR-10 and an IPC of 10 for CIFAR-100, Tiny ImageNet and ImageNet-1K.}    
	\begin{tabular}{c|cccc}  \toprule  $\alpha$ & CIFAR-10 & CIFAR-100 & Tiny ImageNet & ImageNet-1K \\  \midrule  0     & 74.2±0.3 & 44.0±0.2 & 29.1±0.5 & 23.9±0.7 \\    0.1   & 75.0±0.2 & 44.5±0.3 & 30.0±0.3 & 24.8±0.2 \\    1     & 75.0±0.2 & 44.6±0.4 & 30.0±0.2 & 24.4±0.3 \\    10    & 74.8±0.3 & 44.6±0.3 & 29.9±0.2 & 24.3±0.2 \\    \bottomrule
	\end{tabular}%  
\label{tab:ce}%
\end{table}%

\subsection{Comparison of Distance Functions}\label{sec:dist}
In this work, we utilize the \emph{Manhattan distance} (also known as $L_1$ distance) as the default distance function for assessing the logits disagreement between teachers and students. To demonstrate its efficacy, we compare it with two alternative distance metrics, namely, the Euclidean distance and the cosine similarity, across three experimental scenarios including CIFAR-10 with an IPC of 50, CIFAR-100 and Tiny ImageNet with an IPC of 10. The comparison results are presented in Table \ref{tab:distance}. The results demonstrate that utilizing the \emph{Manhattan distance} for evaluating logits disagreement results in superior performance compared to the other two distance metrics across all three experimental cases, thereby validating our selection of the distance function for our formal implementation.

\begin{table}[htbp]  
	\centering
	\small
	\caption{Comparison of test accuracies (\%) between Manhattan distance and two candidates including Euclidean distance and the cosine similarity on CIFAR-10 with an IPC of 50, CIFAR-100 and Tiny ImageNet with an IPC of 10. }    
	\begin{tabular}{c|c|c|c}      \toprule     & CIFAR-10 & CIFAR-100 & Tiny ImageNet \\ \midrule   Cosine & 68.8±0.3  & 36.2±0.4    & 23.8±0.3 \\    Euclidean & 74.1±0.3 & 43.9±0.2    & 29.2±0.4 \\    \textbf{Manhattan} & \textbf{75.0±0.2} & \textbf{44.5±0.3} & \textbf{30.0±0.3} \\   \bottomrule 
	\end{tabular}%  
	\label{tab:distance}%
\end{table}%

\subsection{Hyper-Parameter Setup and Guidelines}\label{app:hyper}
In Table \ref{tab:parameter}, we present detailed information regarding the hyper-parameter configuration used to obtain the ConvNet results (i.e., the proxy model utilized for distillation) depicted in Table \ref{tab:comp}. In accordance with the ablation study outlined in Section \ref{sec:ablation}, we recommend meticulous tuning of the student model learning rate, $\eta$, and the student model update epoch, $E$, when adapting our method to new datasets or models. This is to avoid an overreliance on distilled features of knowledge from nearly converged checkpoint models. Additionally, considering the stability of tuning the coefficient of the cross-entropy regularizer shown in Appendix \ref{app:alpha}, it is advisable to initiate the tuning process with a smaller value of $\alpha$ (e.g., $0.1$).

\begin{table}[htbp]  
	\tiny
	\centering  
	\caption{Hyper-parameters setup for obtaining the results on ConvNet shown in Table \ref{tab:comp}. Reminder: $\gamma$ - synthetic samples learning rate; $\eta$ - student model learning rate; $E$ - proxy student model update epoch number; $B$ - mini-batch size; $\alpha$ - cross-entropy regularizer coefficient.}    
	\begin{tabular}{c|c|ccccccccc}      \toprule    & IPC   & Model & $\gamma$ & $\eta$ & $E$     & Evaluation Epoch & Iteration & \multicolumn{1}{c}{ZCA} & $B$     & $\alpha$ \\   \midrule \multirow{3}[0]{*}{CIFAR-10} & 50    & ConvNet & 1     & 0.02  & 250   & 1000  & 2500  & TRUE  & full  & 0.10 \\          & 500   & ConvNet & 1     & 0.02  & 250   & 1000  & 2500  & \multicolumn{1}{c}{TRUE} & 2500  & 0.10 \\          & 1000  & ConvNet & 1     & 0.02  & 250   & 600   & 2500  & \multicolumn{1}{c}{TRUE} & 2500  & 0.10 \\  \midrule  \multirow{3}[0]{*}{CIFAR-100} & 10    & ConvNet & 1     & 0.02  & 250   & 1000  & 2500  & TRUE  & full  & 0.10 \\          & 50    & ConvNet & 1     & 0.04  & 250   & 1000  & 2500  & TRUE  & 2500  & 0.10 \\          & 100   & ConvNet & 1     & 0.04  & 250   & 600   & 2500  & TRUE  & 2500  & 0.10 \\  \midrule  \multirow{3}[0]{*}{Tiny ImageNet} & 10    & ConvNetD4 & 20    & 0.10   & 250   & 1000  & 2500  & FALSE & 500   & 0.10 \\          & 50    & ConvNetD4 & 40    & 0.10   & 250   & 1000  & 2500  & FALSE & 500   & 0.10 \\          & 100   & ConvNetD4 & 40    & 0.10   & 250   & 600   & 2500  & FALSE & 500   & 0.10 \\   \midrule \multirow{3}[0]{*}{ImageNet-1K} & 10    & ConvNetD4 & 20     & 0.10  & 250   & 1000  & 2500  & FALSE & 500   & 0.10 \\          & 50    & ConvNetD4 & 20    & 0.10  & 250   & 600   & 2500  & FALSE & 500   & 0.10 \\          & 100   & ConvNetD4 & 20    & 0.10  & 250   & 600   & 2500  & FALSE & 500   & 0.10 \\    \bottomrule
	\end{tabular}%  
\label{tab:parameter}%
\end{table}%

\subsection{Visualization Study}\label{app:visual}
In this section, we visualise our synthetic samples for CIFAR-10, CIFAR-100, Tiny ImageNet and ImageNet-1K. Due to the large class numbers of CIFAR-100, Tiny ImageNet and ImageNet-1k, we only exhibit the generated synthetic samples for the former 50 classes. 

\begin{figure}[htbp]
	\begin{center}
		\centerline
		{\includegraphics[width=0.95\textwidth]{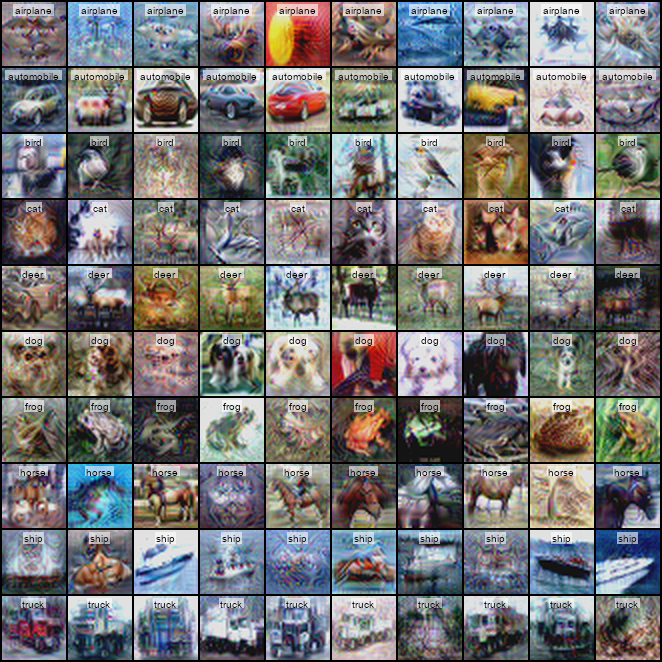}}
		\caption{Synthetic image visualization for CIFAR10.}
%		\label{fig:c10}
	\end{center}
	
\end{figure}

\begin{figure}[htbp]
	\centering
	\begin{minipage}{0.47\textwidth}  % Adjust the width of the minipage as needed
		\centering
		\includegraphics[width=\linewidth]{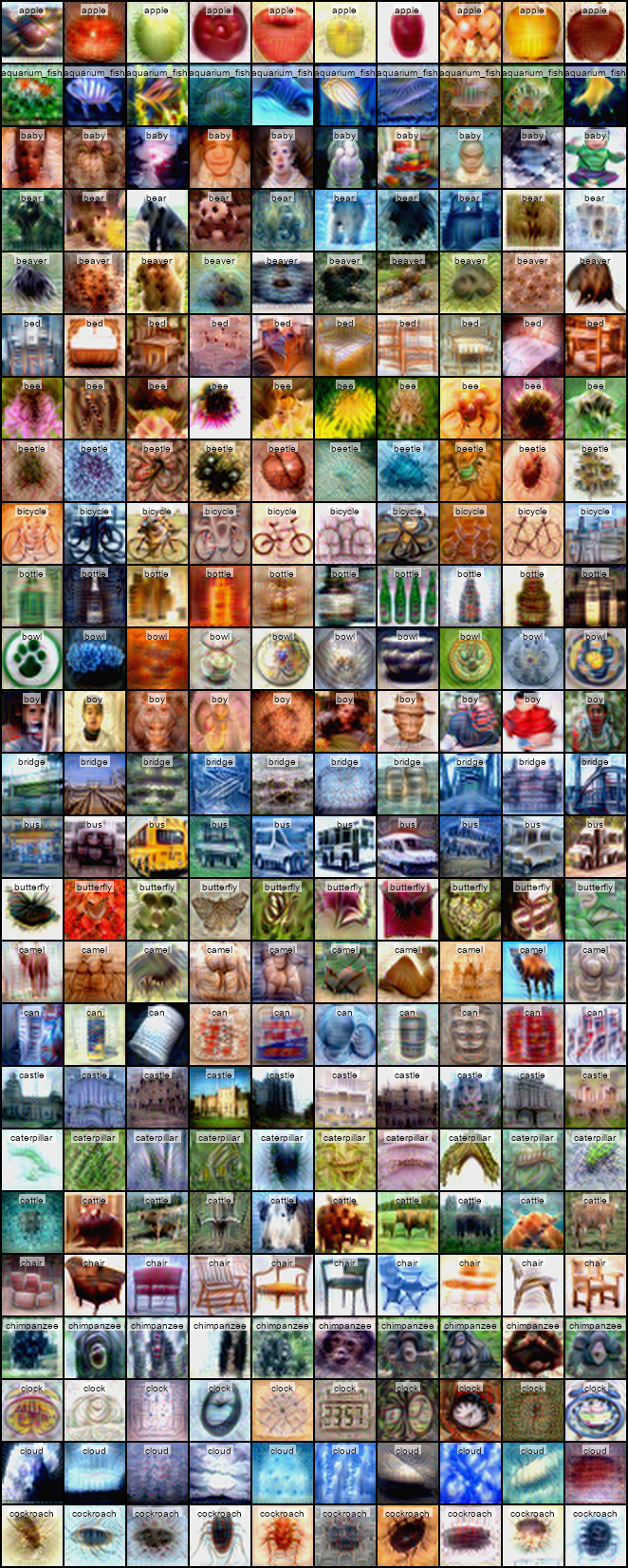} % Adjust the image name and width as needed
		\caption{CIFAR100 - [1,25].}
	\end{minipage}
	\hspace{0.3cm} % To add space between the two figures
	\begin{minipage}{0.47\textwidth}  % Adjust the width of the minipage as needed
		\centering
		\includegraphics[width=\linewidth]{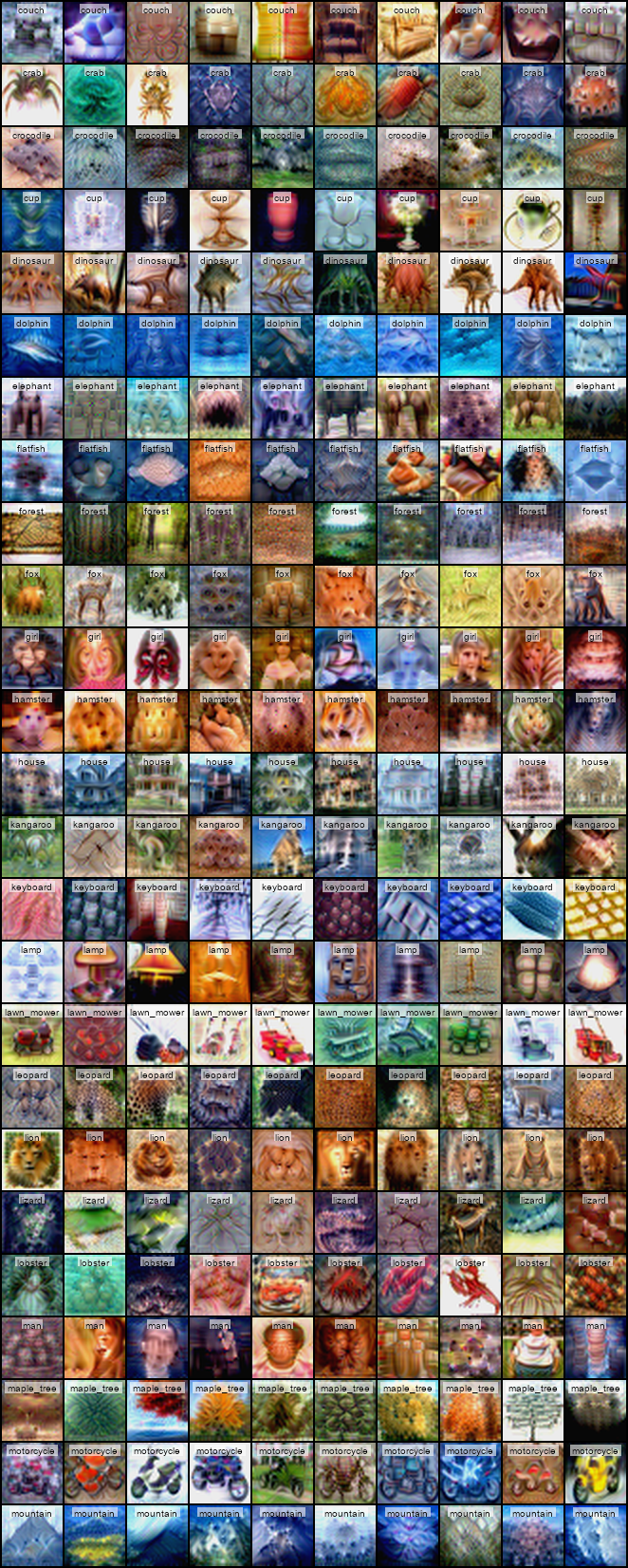} % Adjust the image name and width as needed
		\caption{CIFAR100 - [26,50].}
	\end{minipage}%
%\caption{Synthetic image visualization for CIFAR10 from class 1 to 50.}
\end{figure}

\begin{figure}[htbp]
	\centering
	\begin{minipage}{0.47\textwidth}  % Adjust the width of the minipage as needed
		\centering
		\includegraphics[width=\linewidth]{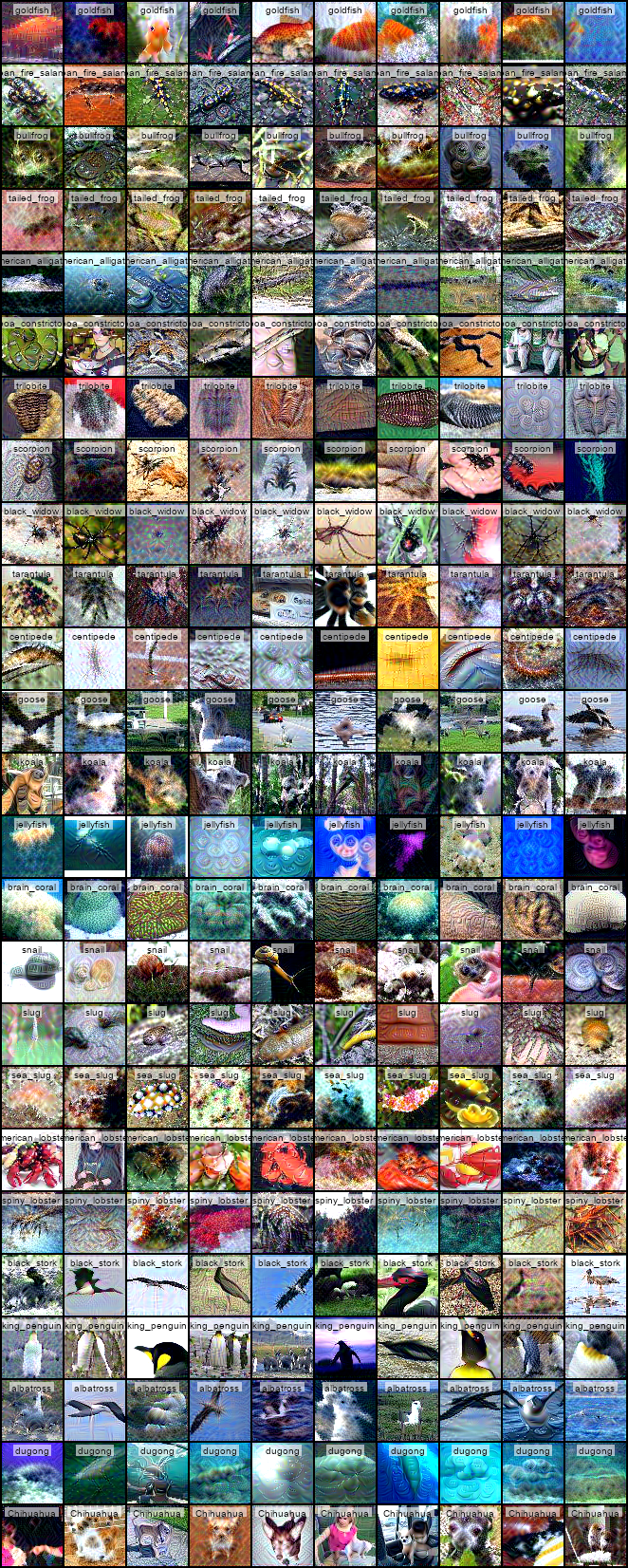} % Adjust the image name and width as needed
		\caption{Tiny ImageNet - [1,25].}
	\end{minipage}
	\hspace{0.3cm} % To add space between the two figures
	\begin{minipage}{0.47\textwidth}  % Adjust the width of the minipage as needed
		\centering
		\includegraphics[width=\linewidth]{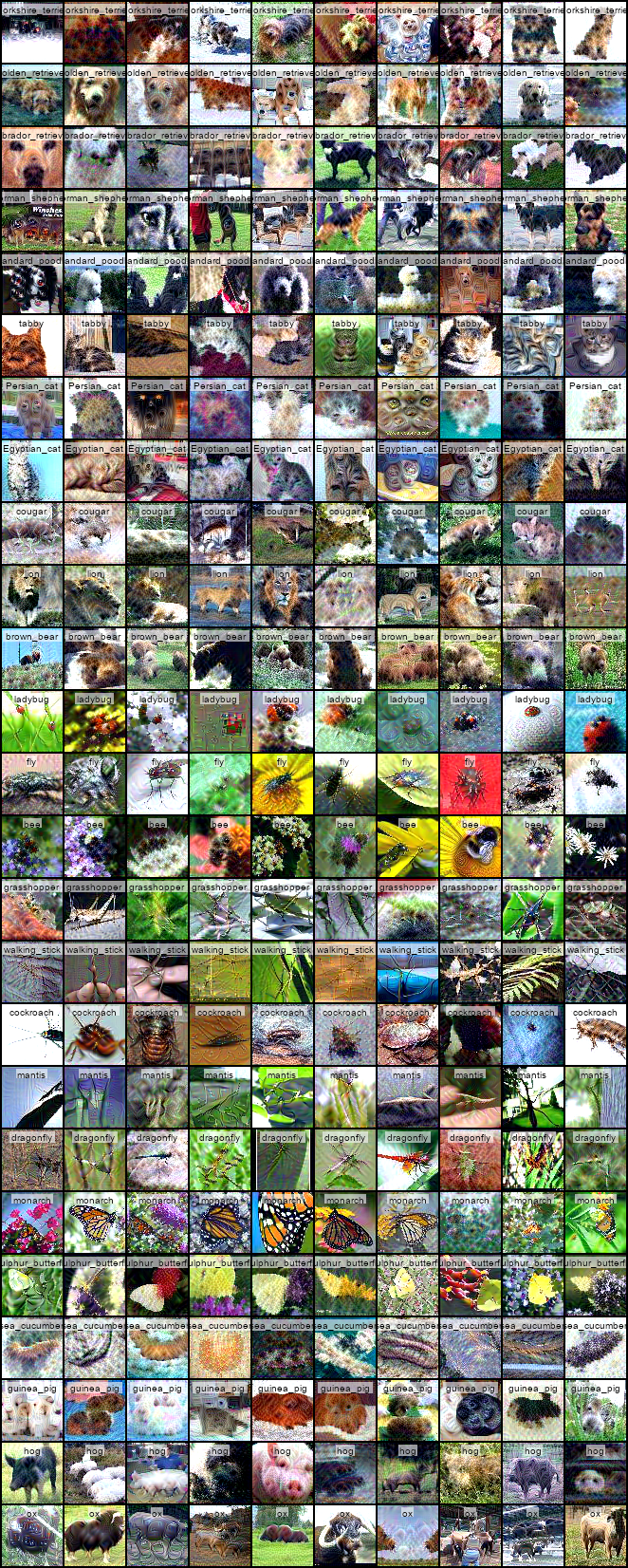} % Adjust the image name and width as needed
		\caption{Tiny ImageNet - [26,50].}
	\end{minipage}%
	%\caption{Synthetic image visualization for CIFAR10 from class 1 to 50.}
\end{figure}

\begin{figure}[htbp]
	\centering
	\begin{minipage}{0.47\textwidth}  % Adjust the width of the minipage as needed
		\centering
		\includegraphics[width=\linewidth]{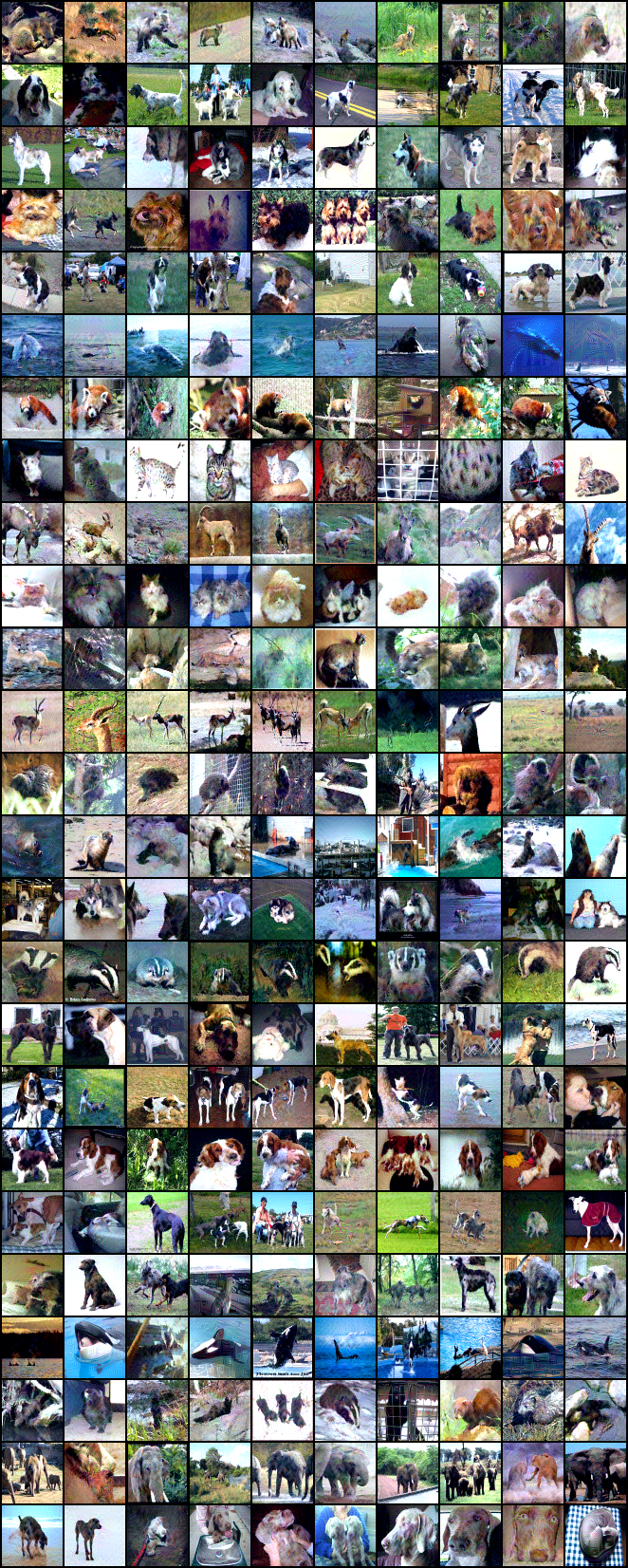} % Adjust the image name and width as needed
		\caption{ImageNet-1K - [1,25].}
	\end{minipage}
	\hspace{0.3cm} % To add space between the two figures
	\begin{minipage}{0.47\textwidth}  % Adjust the width of the minipage as needed
		\centering
		\includegraphics[width=\linewidth]{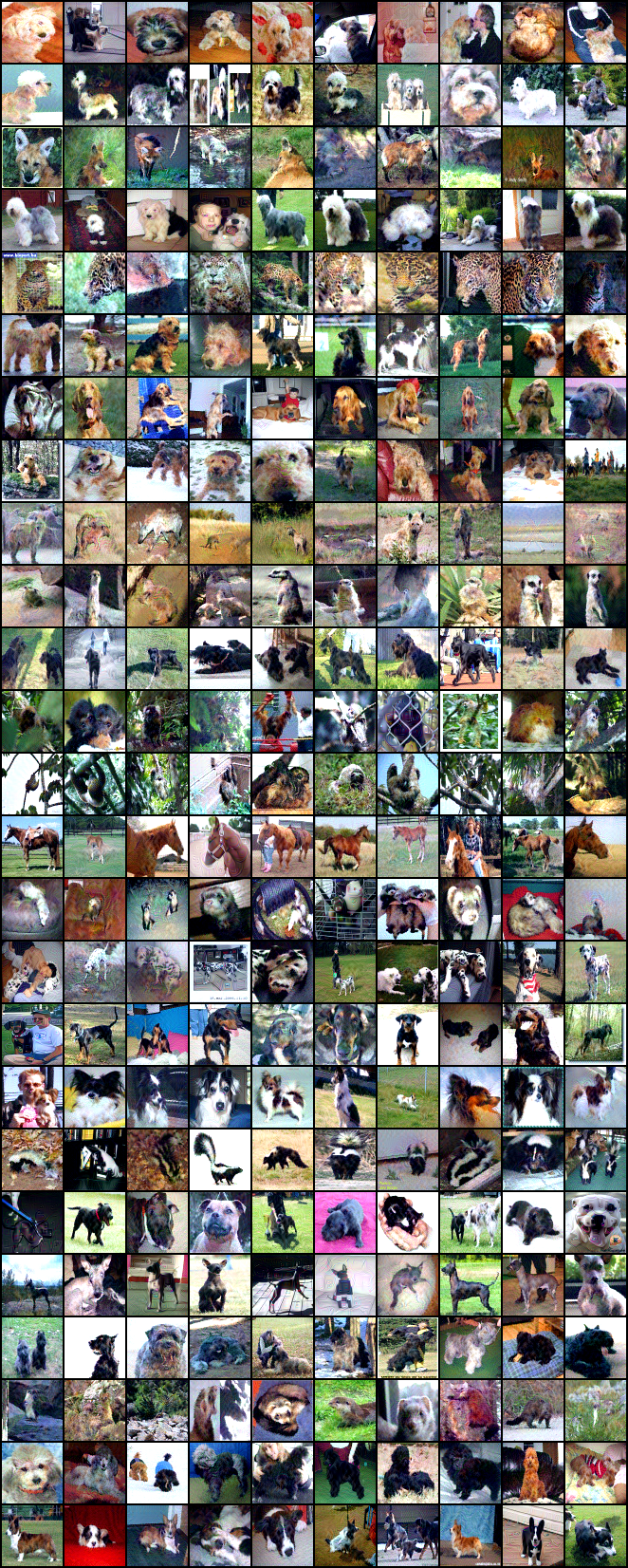} % Adjust the image name and width as needed
		\caption{ImageNet-1K - [26,50].}
	\end{minipage}%
	%\caption{Synthetic image visualization for CIFAR10 from class 1 to 50.}
\end{figure}

\end{document}